\documentclass[sigconf]{acmart}

\usepackage{makecell}
\usepackage{subcaption}
\usepackage{float}

\AtBeginDocument{%
  \providecommand\BibTeX{{%
    \normalfont B\kern-0.5em{\scshape i\kern-0.25em b}\kern-0.8em\TeX}}}

\begin{document}

\newcommand{\ie}{i.e., }
\newcommand{\eg}{e.g., }
\newcommand{\etc}{\textit{etc.}}
\newcommand{\viz}{\textit{viz. }}
\newcommand{\etal}{\textit{et. al.}}

\title{TsSHAP: Robust model agnostic feature-based explainability for univariate time series forecasting}


\author{Vikas C. Raykar}
\authornote{The authors were associated with IBM Research during the work.}
\email{vikasraykar@gmail.com}

\author{Arindam Jati}
\email{arindam.jati@ibm.com}

\author{Sumanta Mukherjee}
\email{sumanm03@in.ibm.com}

\author{Nupur Aggarwal}
\authornotemark[1]
\email{nupur.wal@gmail.com}
\affiliation{
  \institution{IBM Research}
  \city{Bangalore}
  \country{India}
}

\author{Kanthi Sarpatwar}
\authornotemark[1]
\email{sarpatwa@us.ibm.com}

\author{Giridhar Ganapavarapu}
\email{giridhar.ganapavarapu@ibm.com}

\author{Roman Vaculin}
\email{vaculin@us.ibm.com}
\affiliation{\institution{IBM Research}
  \city{Yorktown}
  \country{USA}
 }

\begin{abstract}
A trustworthy machine learning model should be accurate as well as explainable. Understanding why a model makes a certain decision defines the notion of explainability. While various flavors of explainability have been well-studied in supervised learning paradigms like classification and regression, literature on explainability for time series forecasting is relatively scarce. 
In this paper, we propose a feature-based explainability algorithm, TsSHAP, that can explain the forecast of any black-box forecasting model. The method is agnostic of the forecasting model being explained, and can provide explanations for a forecast in terms of interpretable features defined by the user a priori. 
The explanations are in terms of the SHAP values which are obtained by applying the TreeSHAP algorithm on a surrogate model that learns a mapping between the interpretable feature space and the forecast of the black-box model.   
Moreover, we formalize the notion of local, semi-local, and global explanations in the context of time series forecasting, which can be useful in several scenarios.
We validate the efficacy and robustness of TsSHAP through extensive experiments on multiple datasets. 
\end{abstract}



\maketitle

\section{Introduction}
The goal of time series forecasting is to predict the future observations of a temporally ordered sequence from its history. Forecasting is central to many domains, including retail, manufacturing, supply chain, workforce, weather, finance, sensor networks, and health sciences. An accurate forecast is key to successful business planning. There exist several time-series forecasting methods in the literature~\cite{hyndman_book}. They are broadly classified into \textbf{statistical models} (SARIMA, Exponential Smoothing, Prophet, Theta, \etc), modern \textbf{machine learning models} (tree-based ensembles like XGBoost, CatBoost, LightGBM, Gaussian Process Regression, \etc), and more recent \textbf{deep learning models} including DeepAR~\cite{salinas2020deepar}, N-BEATS~\cite{oreshkin2019n}, Informer~\cite{zhou2021informer}, and SCINet~\cite{liu2021time}.

Although the classical statistical models are well-motivated and inherently explainable, recent research demonstrates that modern machine learning and deep learning approaches outperform them in several benchmarks.
For example, in the latest M competition (M5)~\cite{makridakis2022m5}, gradient boosted trees and several combinations of boosted trees and deep neural networks were among the top performers.
In the M4 competition, a hybrid modeling technique between statistical and machine learning models was the winner~\cite{smyl2020hybrid}.
More recently, a pure deep learning model N-BEATS has outperformed the M4 winner's performance~\cite{oreshkin2019n}.
While the more complex machine learning and deep learning techniques outperform the statistical methods, particularly due to the availability of large-scale training datasets; the former models are partially or fully black-box, and hence, it is hard or impossible to explain a certain forecasting prediction made by them.
This engenders a lack of trust in deploying these models for sensitive real-world tasks.

Since performance is an important factor along with explainability, effort in devising novel and complex machine learning models for time series forecasting is actively ongoing. Researchers are recently focusing on building algorithms that can explain the predictions of a complex black-box model.
Although research on explainability is well-matured for supervised classification and regression tasks, the forecasting domain still lacks it (see \S~\ref{sec:Prior art} for details).

\subsection{Our contribution}
In this paper, we propose a model agnostic feature-based explainability method, named TsSHAP, that can explain the prediction of \textit{any} black-box forecasting model in terms of SHAP value-based feature importance scores~\cite{lundberg2017unified}.
The novelties of the proposed method are the following.
\begin{itemize}
    \item TsSHAP can explain the prediction(s) of any black-box forecaster model from a set of interpretable features defined by the user \textit{a priori}. It only needs to access the outputs of the \texttt{fit()} and \texttt{predict()} functions of the model being explained\footnote{Following conventional machine learning terminology, \texttt{fit()} and \texttt{predict()} refer to the training and inference functions of the forecasting algorithm respectively.}. It always needs to access the \texttt{predict()} function, and \texttt{fit()} is required only for a specific set of forecasters (see \S~\ref{subsec:Backtested historical forecasts} for details).
    \item It amalgamates the notion of model surrogacy, and the strength of the fast TreeSHAP algorithm into a suitable and ready-to-use post-hoc explainability method for time series forecasting.
    \item The novel TsSHAP methodology uniquely reduces the surrogate forecasting task into a regression problem where the surrogate model learns a mapping between the interpretable feature space and the output of the forecaster model.
    \item It introduces multiple scopes of explanation, \viz local, semi-local, and global, which can be useful to obtain useful explanations in several real-world scenarios.
\end{itemize}

\section{Background and notations}
\subsection{Time series forecasting}
Let $f(t):\mathbb{Z}\to\mathbb{R}^1$ represent a latent \textbf{univariate} time series for any discrete time index $t \in \mathbb{Z}$. We observe a sequence of historical \emph{noisy} (and potentially missing) values $y(t)$ for $t \in [1,\dots,T]$  such that in expectation $\mathbb{E}[y(t)]=\mathbb{E}[f(t)]$. For example, in the retail domain $y(t)$ could represent the daily sales of a product and $f(t)$ the true latent demand for the product.

The task of \textbf{time series forecasting} is to estimate $f(t)$ for all $t >T$ based on the observed historical time series $y(t)$ for $t \in [1,\dots,T]$. The time series forecast is typically done for a fixed number of periods in the future, referred to as the \textbf{forecast horizon}, $H$. 
We denote
\begin{equation}
    \hat{f}\left(T+h|y(1),...,y(T)\right) \stackrel{\triangle}{=} \hat{f}\left(T+h|T\right)\;\text{for}\:h \in [1,\dots,H]
\end{equation}
as the projected forecast over time horizon $H$ based on the historical observed time series $y(1),...,y(T)$ of length $T$. 

\subsubsection{Forecasting with external regressors}
In many domains, the value of the target time series potentially depends on several other external time series, called \textbf{related external regressors}. For example, in the retail domain, the sales are potentially influenced by discounts, promotions, events, and weather. 
Let $z_1(t),\hdots,z_{K}(t)$ be the $K$ external regressors available that can potentially influence the target time series $f(t)$. Let\footnote{Equation {\ref{eqn:exog}} assumes that the future value of the external regressor is also available, which is true in many situations.}
\begin{equation}  \hat{f}\left(T+h|y(1),\hdots,y(T),\left(z_k(1),\hdots,z_k(T+h)\right)_{k=1}^{K}\right)\:\text{for }\:h \in [1,\dots,H]
\label{eqn:exog}
\end{equation}
be the forecasted time series for a forecast horizon $H$ based on the historical observed time series $y(1),\hdots,y(T)$ of length $T$ and the $K$ external regressors $\left(z_k(1),\hdots,z_k(T+h)\right)_{k=1}^{K}$ each of length $T+h$. 

\subsection{Explainability}\label{subsec:Explainability}
Explainability is the degree to which a human can understand the cause of a decision (or prediction) made by a prediction model~\cite{miller_ai_2019}.
An explanation can be based on an instance, a set of features, or (for a time series) inherent components (\eg trend, seasonality \etc).
In this paper, our focus is feature-based explanation.

A feature-based explanation $\phi \in \mathbb{R}^d$ is a $d$-dimensional vector where each element corresponds to the importance /contribution score for the corresponding feature. 
Several feature-based explanation methods have been proposed in the supervised classification and regression literature (see \S~\ref{sec:Prior art}). We will adopt SHAP for explanation because of it's intuitive nature~\cite{lundberg2017unified} and state-of-the-art performance~\cite{treeshap_paper}. 

\subsubsection{SHapley Additive exPlanations (SHAP)}
SHAP is a game theoretic approach to explain the output of any feature-based machine learning model.
It connects optimal credit allocation with local explanations using the Shapley values from game theory and their related extensions. 
SHAP computes the contribution of a feature towards the conditional expectation of the model's output $g_x(S)=\mathbb{E}(g(x)|\text{do}(X_S=x_S))$, by estimating the expected difference in the predicted result in the absence of the corresponding feature~\cite{treeshap_paper}\footnote{We follow causal do-notation here~\cite{treeshap_paper}.}.
\begin{equation}
    \phi_i(g,x)=\sum_{R \in \mathcal{R}} \frac{1}{M!} \left[
    g_x\left( P_i^R \cup i \right) - g_x\left( P_i^R \right)
    \right]
\end{equation}
where, $\mathcal{R}$ is the set of all feature orderings, M is the number of features, $P_i^R$ is the subset of feature that come before feature $i$ in ordering $R$. In our scenario, $g(x)$ will denote the surrogate model (see \S~\ref{subsec:Surrogate model}).

\section{Prior art}
\label{sec:Prior art}
Several research efforts have been devoted toward formulating model agnostic explainability for classification and regression methods that work with image and tabular data.
Recent developments include LIME~\cite{ribeiro2016should}, DeepLIFT~\cite{shrikumar2017learning}, Layer-Wise Relevance Propagation (LRP)~\cite{bach2015pixel}, Classical Shapley values~\cite{lipovetsky2001analysis}, SHAP~\cite{lundberg2017unified} and its several variants like KernelSHAP~\cite{lundberg2017unified}, TreeSHAP~\cite{treeshap_paper}, and DeepSHAP~\cite{lundberg2017unified}.
A detailed survey can be found in~\cite{molnar_book_2019}.

Research on explainability for time series data mainly started with time series classification problems.
Mokhtari \etal~\cite{mokhtari2019interpreting} employed SHAP to derive global feature-based explanations for several time series classifiers.
Madhikermi \etal~\cite{madhikermi2019explainable} utilized LIME for explaining the predictions of SVM and neural network classifiers designed to detect faults in an air handling unit.
Schlegel \etal~\cite{schlegel2019towards} performed an extensive evaluation of multiple explainers like LIME, LRP, and SHAP to interpret multiple binary and multi-class time series classifiers.
The authors found SHAP to be the most robust model agnostic explainer, while the other algorithms tend to be biased towards specific model architectures.
Assaf \etal~\cite{assaf2019explainable} proposed a white-box explainability method to generate two-dimensional saliency maps for time series classification.

The researchers in~\cite{garcia2020shapley} employed SHAP to analyze a neural network-based forecasting model that predicts the nitrogen dioxide concentration in a city.
However, the local and global SHAP explanations were not evaluated.
Saluja \etal~\cite{saluja2021towards} demonstrated the explanations derived with SHAP and LIME for a sales time series forecasting model. 
The work also performed a human study to find the usefulness of the explanations.
However, the work of~\cite{saluja2021towards} employed only a support vector regressor as the forecasting model, while the proposed TsSHAP method supports any forecasting model and it only needs to access the \texttt{fit()} and \texttt{predict()} methods of the model (we experiment with six different forecasting models, see \S~\ref{sec:Experiments}).
The tree-based surrogate modeling of TsSHAP helps to achieve this.
Moreover, we evaluate TsSHAP on five different datasets as compared to~\cite{saluja2021towards} which focused on a specific case study.
Furthermore, we incorporate a more generic and much larger set of \textit{interpretable features} that is well-suited for time series forecasting. 
Recently, Rajapaksha \etal~\cite{DBLP:journals/corr/abs-2111-07001} proposed the LoMEF framework to provide local explanations for global time series forecasting models by training explainable statistical models at the local level.
Our proposed method is inherently different than~\cite{DBLP:journals/corr/abs-2111-07001} since the explanations are in terms of the interpretable features that the user can define \textit{a priori}. We provide explanations in terms of the SHAP values which satisfy all three desirable properties (local accuracy, missingness, and consistency) of additive feature attribution methods, and thus, was found to be have more consistency with human intuition~\cite{lundberg2017unified}. Moreover, the TsSHAP can provide explanations at multiple scopes \viz local, semi-local, and global.

\section{TsSHAP Methodology}

\subsection{Surrogate model}\label{subsec:Surrogate model}
\begin{figure}[t]
\centering
\includegraphics[width=\columnwidth]{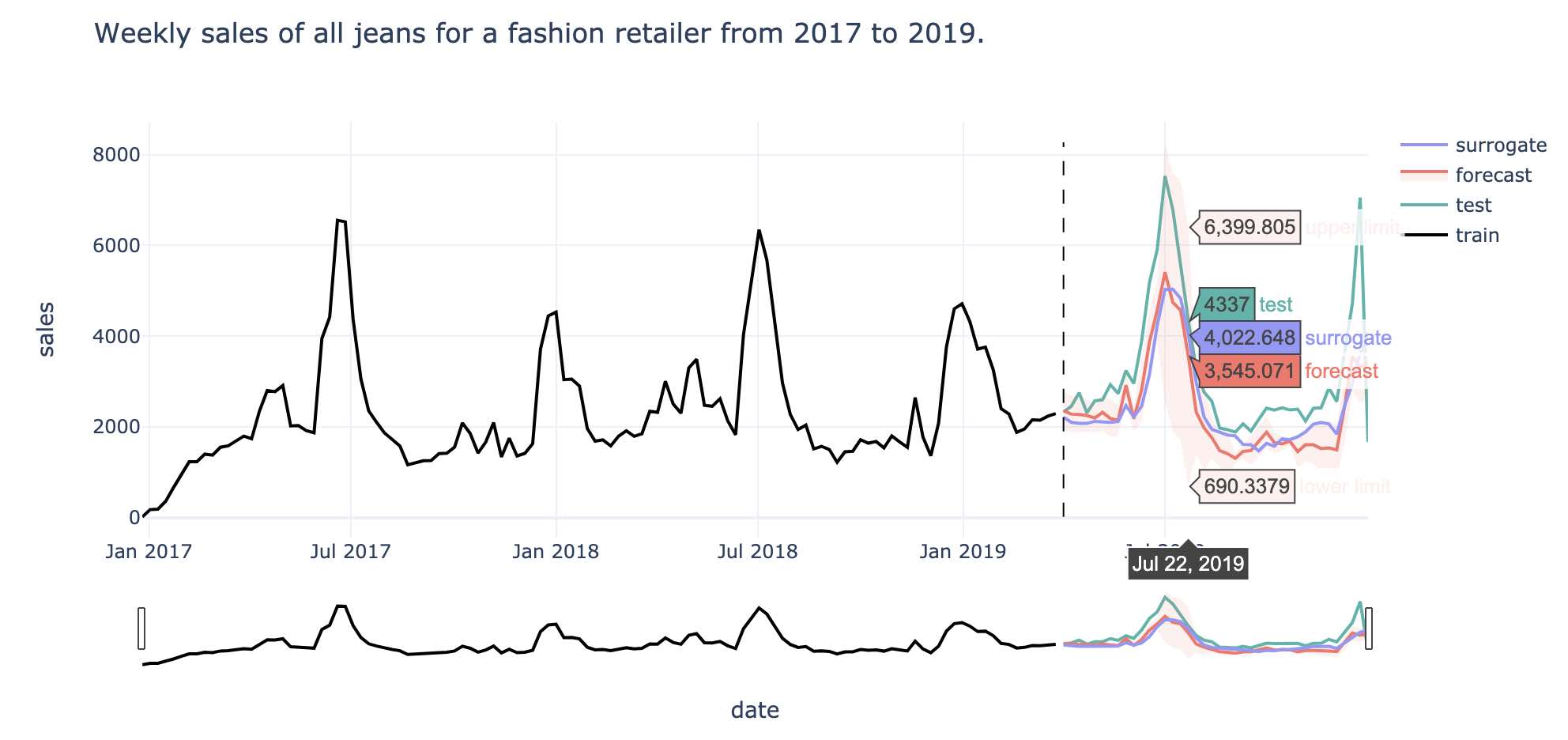}
\caption{The forecasts from the surrogate model and the forecasts from the forecaster. The surrogate model is trained to mimic the forecasts of the corresponding forecaster.}
\label{fig:timeshap_surrogate}
\end{figure}

TsSHAP is model agnostic. Since we do not have access to the actual internals of the forecaster model, we first learn a surrogate model $g$. 
\begin{equation}
g(T+h|T)=g\left(T+h|y(1),...,y(T)\right)\:\text{for}\:h=1,\hdots,H.
\end{equation}
The surrogate model $g$ produces a point-wise forecast approximation \textbf{of the forecaster}. While the original forecaster learns to predict $y(T+h)$ based on $y(1),\hdots,y(T)$ the surrogate model is trained to predict the forecasts $\hat{f}(T+h)$ made by the forecaster. Essentially we want to mimic the forecaster using a surrogate model since we aim to explain the forecasts made by the forecaster and not the original time series. Figure~\ref{fig:timeshap_surrogate} illustrates this concept with an example time series from a real-world dataset. 

\subsection{Backtested historical forecasts}\label{subsec:Backtested historical forecasts}

\begin{figure}[t]
\centering
\includegraphics[width=.9\columnwidth]{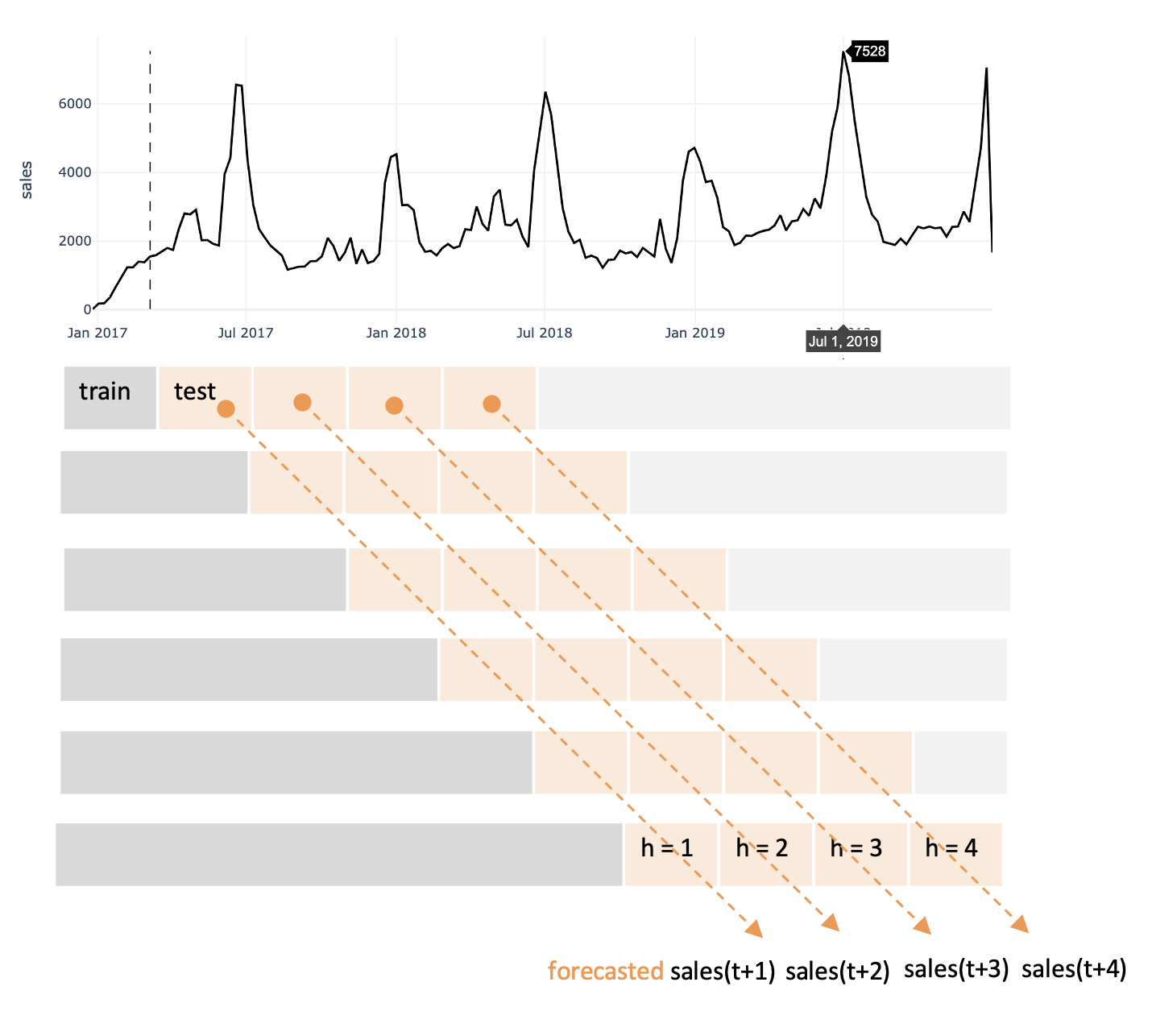}
\caption{Backtested historical forecasts using an expanding window splitter to train the surrogate forecasting model.}
\label{fig:timeshap_backtest}
\end{figure}

To generate the training data for the surrogate model, we use \textbf{backtesting}. For a given univariate time-series,  we produce a sequence of train and test partitions using an \emph{expanding window} strategy (see Figure~\ref{fig:timeshap_backtest}).  The expanding window partitioning strategy uses more and more training data while keeping the test window size fixed to the forecast horizon. The forecaster is trained (\texttt{fit()} is accessed) on the train partition and evaluated (\texttt{predict()} is accessed) on the test split. All the test split predictions are concatenated to get the backtested historical forecasts for each step of the forecast horizon. 

Note that we need to access the \texttt{fit()} function only for a set of classical forecasters like SARIMA or exponential smoothing, where we train the forecaster for every expanding window. 
For most of the machine learning and deep learning forecaster models, we do not need to access their \texttt{fit()} functions because a single model is generally trained (beforehand) on the entire data.  


\subsection{Regressor reduction}
We now have an original time series and the corresponding backtested forecast time series for each step of the forecast horizon. The goal of the surrogate model is to learn to predict the backtested forecast time series based on the original time series. We construct a surrogate time series forecasting task by reducing it to a supervised regression problem.
A standard supervised regression task takes a $d$-dimensional feature vector as input ($\mathbf{x} \in \mathbb{R}^d$) and predicts a scalar $y \in \mathbb{R}$. The regressor $y = f(\mathbf{x})$ is learnt based on a labelled training dataset $\left(\mathbf{x}_i,y_i\right)$, for $i=1,\hdots,n$.  The input time series sequence does not satisfy classical feature definitions.  
Instead, we process the original time series to compute various interpretable features.

\subsection{Interpretable features}
For each time point $t$, we compute features $\mathbf{x}(t) \in \mathbb{R}^d$ based on the original time series values observed so far. 
The surrogate model uses these $d$ features to predict the backtested forecast for each step in the forecast horizon. 
We focus only on \textbf{interpretable features} such that the explanations produced on the features are easily comprehensible by the end-user. The set of interpretable features can be defined by the (expert) user as well.
As a starting point, we define a set of seven feature families, viz., lag, seasonal lag, rolling window, expanding window, date and time encoding, holidays, and trend-related features. Table~\ref{tab:features-1} and \ref{tab:features-2} list the features.





\begin{table}[t]
\centering
\small
\caption{Interpretable features based on lag, seasonal lag, expanding and rolling window and trend based features. An example is provided with each feature description.}
\begin{tabular}{lp{0.45\columnwidth}}
\toprule
\textbf{feature name}  & \textbf{feature description}\\
\midrule
\multicolumn{2}{l}{\texttt{LagFeatures(lags=3)}}\\
\multicolumn{2}{l}{Value at previous time steps.}\\
\texttt{sales(t-3)} & The value of the time series at (t-3) previous time step. \\ 
 \midrule
 \multicolumn{2}{l}{\texttt{SeasonalLagFeatures(lags=2, m=365)}}\\
 \multicolumn{2}{l}{Value at time steps for the previous seasons.}\\
\texttt{sales(t-2*365)} & The value of time series at (t-2*365) previous time step. \\ 
\midrule
 \multicolumn{2}{l}{\texttt{RollingWindowFeatures(window=3)}}\\
 \multicolumn{2}{l}{Rolling window statistics (mean,max,min).}\\
\texttt{sales-max(t-1,t-3)} & The max of the past 3 values. \\
\midrule
 \multicolumn{2}{l}{\texttt{ExpandingWindowFeatures()}}\\
 \multicolumn{2}{l}{Expanding window statistics (mean,max,min).}\\
\texttt{sales-mean(0,t-1)} & The mean of all the values so far. \\
\midrule
 \multicolumn{2}{l}{\texttt{TrendFeatures(degree=2)}}\\
 \multicolumn{2}{l}{Features to model simple polynomial trend.}\\ 
\texttt{t2} & Feature to model polynomial (of degree 2) trend.\\
\bottomrule
\end{tabular}
\label{tab:features-1}
\end{table}

\begin{table}[t]
\centering
\small
\caption{Interpretable features based on date and time encoding and holidays. Some features like \texttt{fashion-season} might be more relevant for a fashion sales time series forecasting.}
\begin{tabular}{lp{0.65\columnwidth}}
\toprule
\textbf{feature name}  & \textbf{feature description}\\
\midrule
\multicolumn{2}{l}{\texttt{DateFeatures()}}\\
\multicolumn{2}{l}{Date related features.}\\
\texttt{month} &	The month name from January to December.\\
\texttt{day-of-year} &	The ordinal day of the year from 1 to 365.\\
\texttt{day-of-month} &	The ordinal day of the month from 1 to 31.\\
\texttt{week-of-year} &	The ordinal week of the year from 1 to 52.\\
\texttt{week-of-month} &	The ordinal week of the month from 1 to 4.\\
\texttt{day-of-week} &	The day of the week from Monday to Sunday.\\
\texttt{is-weekend} &	Indicates whether the date is a weekend or not.\\
\texttt{quarter} &	The ordinal quarter of the date from 1 to 4.\\
\texttt{season} &	The season Spring/Summer/Fall/Winter.\\
\texttt{fashion-season} &	The fashion season Spring/Summer(January to June) or Fall/Winter(July to December).\\
\texttt{is-month-start} &	Whether the date is the first day of the month.\\
\texttt{is-month-end} &	Whether the date is the last day of the month.\\
\texttt{is-quarter-start} &	Whether the date is the first day of the quarter.\\
\texttt{is-quarter-end} &	Whether the date is the last day of the quarter.\\
\texttt{is-year-start} & Whether the date is the first day of the year.\\
\texttt{is-year-end} &	Whether the date is the last day of the year.\\
\texttt{is-leap-year} &	Whether the date belongs to a leap year.\\
\midrule
\multicolumn{2}{l}{\texttt{TimeFeatures()}}\\
\multicolumn{2}{l}{Time related features.}\\
\texttt{hour} &	The hours of the day.\\
\texttt{minute} &	The minutes of the hour.\\
\texttt{second} &	The seconds of the minute.\\
\midrule
\multicolumn{2}{l}{\texttt{HolidayFeatures(country="IN",buffer=2)}}\\
\multicolumn{2}{l}{Encode country specific holidays as features.}\\
\texttt{holiday-IN} & Indicates whether the date is a IN holiday or not. \\
\bottomrule
\end{tabular}
\label{tab:features-2}
\end{table}

\subsection{Multi-step forecasting}
Recall that we have $H$ backtested time series forecasts which are to be used as targets to learn the surrogate regressor model.
A \textbf{single regressor} is trained for a one-step-ahead forecast horizon and then called recursively to predict multiple steps. Let, $\mathcal{G}(y(1),...,y(T))$ represent the one-step-ahead forecast by the surrogate model trained on the training data. The surrogate produces a one-step-ahead forecast using features based on the time-series till $T$, that is, $y(1),\hdots,y(T)$. The final forecasts from the surrogate model are made recursively as follows,
\begin{equation}
    g(T+1|T) = \mathcal{G}\left(y(1),\hdots,y(T)\right).
\end{equation}
For $h=2,3,\hdots,H$
\begin{equation}
g(T+h|T) = \mathcal{G}\left(y(1),\hdots,y(T),g(T+1|T),\hdots,g(T+h-1|T)\right).
\end{equation}

\subsection{Tree ensemble regressor and SHAP explanation}
In this work, we have used a tree-based ensemble regression model, XGBoost~\cite{xgboost_paper}\footnote{\url{https://github.com/dmlc/xgboost}}, as the surrogate model\footnote{Other models like CatBoost~\cite{catboost_paper} and LightGBM~\cite{lightgbm_paper} can also be employed.}. The selection of tree-based model is biased by its reasonably accurate forecast capability and, the support of the fast (polynomial time) TreeSHAP algorithm which we rely on for the various TsSHAP explanations as described below.~\cite{treeshap_paper}\footnote{\url{https://github.com/slundberg/shap}}. 


\section{TsSHAP explanations}
TsSHAP provides a wide range of explanations based on SHAP values.

\subsection{Scope of explanations} 
We introduce the notion of \textbf{scope of explanation} in the context of time series forecasting.
An explanation answers to either a \textbf{why}-question or a \textbf{what if}-scenario~\cite{molnar_book_2019}. 
We generalize three scopes of explanations.
\begin{enumerate}
    \item A \textbf{local explanation} explains the forecast made by the forecaster at a certain point in time. For example,
    \begin{itemize}
        \item \emph{Why is the sale forecast for July 1, 2019 much higher than the average sales?}
        \item \emph{What is the impact on the sale forecast for July 1, 2019 if I offer a discount of 60\% that week?}
    \end{itemize}
    \item A \textbf{global explanation} explains the forecaster trained on the historical time series.
    \begin{itemize}
        \item \emph{What are the most important attributes the forecaster algorithms rely on to make the forecast?}
        \item \emph{What is the overall impact of discount on the sales forecast?}
    \end{itemize}
    \item A \textbf{semi-local explanation} explains the overall forecast made by a forecaster in a certain time interval.
    \begin{itemize}
        \item \emph{Why is the sales forecast over the next 4 weeks much higher?}
        \item \emph{What is the impact of a particular markdown schedule on a 4 week planning horizon sales forecast?} 
    \end{itemize}
\end{enumerate}
Notions of local and global explanations exist in the supervised learning paradigms. In addition, we introduce the concept of semi-local explanations in the context of time series forecasting, which returns one (semi-local) explanation aggregated over many successive time steps in the forecast horizon. TsSHAP is able to provide explanations suitable for all the scopes.

\subsection{Type of explantions}
For each of the scopes, TsSHAP provides multiple types of explanation.
\subsubsection{For local explanation} TsSHAP provides the following explanations.
\begin{enumerate}
    \item \textbf{SHAP explanation} shows features contributing to push up or down a certain forecast from the base value (the average) to the forecaster model output (as explained in \S~\ref{subsec:Explainability}). Since this is a local explanation, the explanation for the forecast at a different forecast horizon can be quite different.
    \item \textbf{Partial dependence plot~(PDP)} for a given feature shows how the actual forecast (from the surrogate model) varies as the feature value changes.
    \item \textbf{SHAP dependence plot~(SDP)} for a given feature shows how the corresponding SHAP value varies as the feature value changes. SDP is an example of a \textbf{what-if} explanation.
\end{enumerate}

\subsubsection{For global explanation} TsSHAP provides the following explanations.
\begin{enumerate}
    \item \textbf{SHAP explanation} for a global scope means the absolute value of the SHAP scores for each feature across the entire dataset. 
    \item \textbf{Partial dependence plot~(PDP)} in a global scope shows how the average prediction in your dataset changes when a particular feature is changed.
    \item \textbf{SHAP dependence plot~(SDP)} for each feature shows the mean SHAP value for a particular feature across the entire dataset. This shows how the model depends on the given feature, and is a richer extension of the classical PDP.
\end{enumerate}

\subsubsection{For semi-local explanation} TsSHAP provides SHAP feature importance, PDP, and SDP aggregated over a certain time interval in the forecast horizon.

\begin{table}[t]
\centering
\small
\caption{\emph{Datasets} Univariate time series forecasting datasets (with available external regressors) used for experimental validation.}
\begin{tabular}{lllp{0.3\columnwidth}}
\toprule
\textbf{dataset} & \textbf{periodicity} & \textbf{size} & \textbf{external regressors} \\ 
\midrule
\textbf{jeans-sales-daily} & daily & 1095  & discount \\
\multicolumn{4}{p{1.0\columnwidth}}{\small{Daily sales of all jeans for a fashion retailer from 2017 to 2019.}}\\
\midrule
\textbf{jeans-sales-weekly} & weekly & 158  & discount \\
\multicolumn{4}{p{1.0\columnwidth}}{\small{Weekly sales of all jeans for a fashion retailer from 2017 to 2019.}}\\
\midrule
\textbf{us-unemployment} & monthly & 872  & N/A \\
\multicolumn{4}{p{1.0\columnwidth}}{\small{Monthly US unemployment rate since 1948.}}\\
\midrule
\textbf{peyton-manning} & daily & 2905  & N/A \\
\multicolumn{4}{p{1.0\columnwidth}}{\small{Log of daily page views for Wikipedia page of Peyton Manning.}}\\
\midrule
\textbf{bike-sharing} & daily & 731 & \makecell[l]{weather,\\temperature,\\humidity,\\wind speed} \\
\multicolumn{4}{p{1.0\columnwidth}}{\small{Daily count of rental bikes from 2011 to 2012 in Capital bikeshare system.}}\\
\bottomrule
\end{tabular}
\label{tab:datasets}
\end{table}

\begin{table}[t]
\centering
\small
\caption{\emph{Forecasters} Univariate time series forecasting algorithms used for experimental validation. Only Prophet and XGBoost can handle external regressors. Note that some of the forecasters are easily interpretable to validate the correctness of the output of TsSHAP.}
\begin{tabular}{lp{.6\columnwidth}}
\toprule
\textbf{forecaster} & \textbf{description} \\ 
\midrule
\textbf{Naive~\cite{hyndman_book}} & \small{The forecast is the value of last observation.} \\
\textbf{SeasonalNaive~\cite{hyndman_book}} & \small{The forecast is the value of the last observation from the same season of the year.}\\
\textbf{MovingAverage~\cite{hyndman_book}} & \small{A moving average forecast of order $k$, or, $MA(k)$, is the mean of the last $k$ observations of the time series.} \\
\textbf{\makecell[l]{Exponential\\Smoothing~\cite{hyndman_book}}} & \small{The forecast is the exponentially weighted average of its past values. It can be interpreted as a weighted average between the most recent observation and the previous forecast.} \\
\textbf{Prophet~\cite{taylor2018forecasting}} & \small{Prophet is a procedure for forecasting time series data based on an additive model where non-linear trends are fit with yearly, weekly, and daily seasonality, plus holiday effects.} \\
\textbf{XGBoost~\cite{xgboost_paper}} & \small{A machine learning based forecasting to regression reduction algorithm using XGBoost.} \\
\bottomrule
\end{tabular}
\label{tab:forecasters}
\end{table}

\section{Experiments}\label{sec:Experiments}

\subsection{Datasets}
We experimentally validate the proposed explainer algorithm and the corresponding various explanations on five univariate time series forecasting datasets (with any available external regressors) listed in Table~\ref{tab:datasets}. For all datasets, we use a forecast horizon of 10\% of the actual data. Availability of data is provided in the Reprodicibility section (\S~\ref{sec:Reproducibility}).

\subsection{Forecasters}
In order to validate the explanations we present results on six different univariate forecasting algorithms listed in Table~\ref{tab:forecasters}. Only some forecasters (Prophet~\cite{taylor2018forecasting} and XGBoost~\cite{xgboost_paper}) can explicitly incorporate external regressors.

\subsection{Robustness evaluation via time series perturbation}\label{subsec:perturber}
Robustness of TsSHAP explanations is evaluated by perturbing the original time series (in \S~\ref{sec:metrics}, we will describe the evaluation metrics that will use the perturbed time series).
To generate bootstrapped samples (\ie multiple perturbed versions) of a time series $y(t)$, we employ the \textbf{block bootstrap} algorithm~\cite{buhlmann2002bootstraps,kreiss2012bootstrap}.
The block bootstrap algorithm extends Efron's bootstrap~\cite{efron1992bootstrap} setup for independent observations to time series where the observations ($y(t)$ for different values of $t$) might not be independent.

To preserve the trend-cycle of the time series~\cite{hyndman_book}, we first decompose $y(t)$ into trend-cycle and residual components using a moving average model of order $m$,
\begin{align}
    y_{\text{trend-cycle}}(t) &= \frac{1}{m} \sum_{j=-k}^{j=k} y(t+j)
    \quad \text{where,} \quad m=2k+1 \\
    y_{\text{residual}}(t) &= y(t) - y_{\text{trend-cycle}}(t).
\end{align}
The residual series $y_{\text{residual}}(t)$ is then utilized by the block bootstrap algorithm. 
It samples contiguous blocks of $y_{\text{residual}}(t)$ at random (with replacement) and concatenates them together to produce a bootstrapped residual,
\begin{equation}
    \tilde{y}_{\text{residual}}(t) = \texttt{Block-Bootstrap}\left(y_{\text{residual}}(t), L_b\right),
\end{equation}
where, $L_b$ is the block-length. The final perturbed series is then obtained by summing the bootstrapped residual to the trend-cycle component of the original series,
\begin{equation}
    \tilde{y}(t) = y_{\text{trend-cycle}}(t) + \tilde{y}_{\text{residual}}(t).
\end{equation}

\subsection{Metrics to evaluate explanations}\label{sec:metrics}
We have extended three well-adopted evaluation metrics in the context of time series, viz. faithfulness, sensitivity, and complexity. Faithfulness measures the consistency of an explanation, sensitivity measures the robustness, and complexity measures the sparsity of an explanation~\cite{ijcai2020-417}.

\subsubsection{Faithfulness}
It is high when the change in predicted values due to the change in feature values is correlated to the change in the feature importance in the explanations. 
Let $\hat{f}_X(T+h|T)$ represents predictions for $\:h=1,\hdots,H$ when $\hat{f}$ was trained on data $X$. Let $\mathcal{N}(X)$ represents a neighbourhood around $X$, obtained through perturbation (\S~\ref{subsec:perturber}).
$\phi_{j}(\hat{f}(T+h|T)$ represents the feature importance for the $j^{th}$ feature. Then faithfulness is given by  
\begin{equation}
    \mu_{F} = \text{correlation} ( \delta_{\hat{f}} , \delta_{\phi})
\end{equation}
where
\begin{align*}
    \delta_{\hat{f}} &= \hat{f}_X(T+h|T) - \hat{f}_{\mathcal{N}_X}(T+h|T) \\
    \delta_{\phi} &= \sum_j (\phi_j(\hat{f}_{X}(T+h|T)) - {\phi_j(\hat{f}_{\mathcal{N}_X}}(T+h|T))
\end{align*}
A higher value of faithfulness is preferred in an explanation.

\subsubsection{Sensitivity}
It measures the change in explanations when model inputs and outputs vary by an insignificant amount. A lower value of sensitivity is preferred in an explanation. 
We define average sensitivity as follows.
\begin{equation}
    \mu_{S} = \int_{z} \mathcal{D}(\phi(f_X(x)) , \phi(f_{\mathcal{N}_X}(z)) ) \mathbb{P}_x(z) dz
\end{equation}
where $\mathcal{D}$ is a distance measure (Euclidean in this paper) and $z$ is a neighborhood of  $x$.

\subsubsection{Complexity}
A simple explanation is easy to understand. 
To measure the complexity of an explanation, we calculate the entropy of the fractional distribution of feature importance scores. 
\begin{equation}
    \mu_C = -\sum_{i=1}^{d}p_{i} \ln{p_{i}}
\end{equation}
where, $p_i = {\phi_{i}}/{\sum_{j=1}^{d}{\phi_{j}}}$ and $d$ denotes the total number of features. 

\section{Results}\label{sec:Results}

\begin{figure}[t]
\centering
\includegraphics[width=\columnwidth]{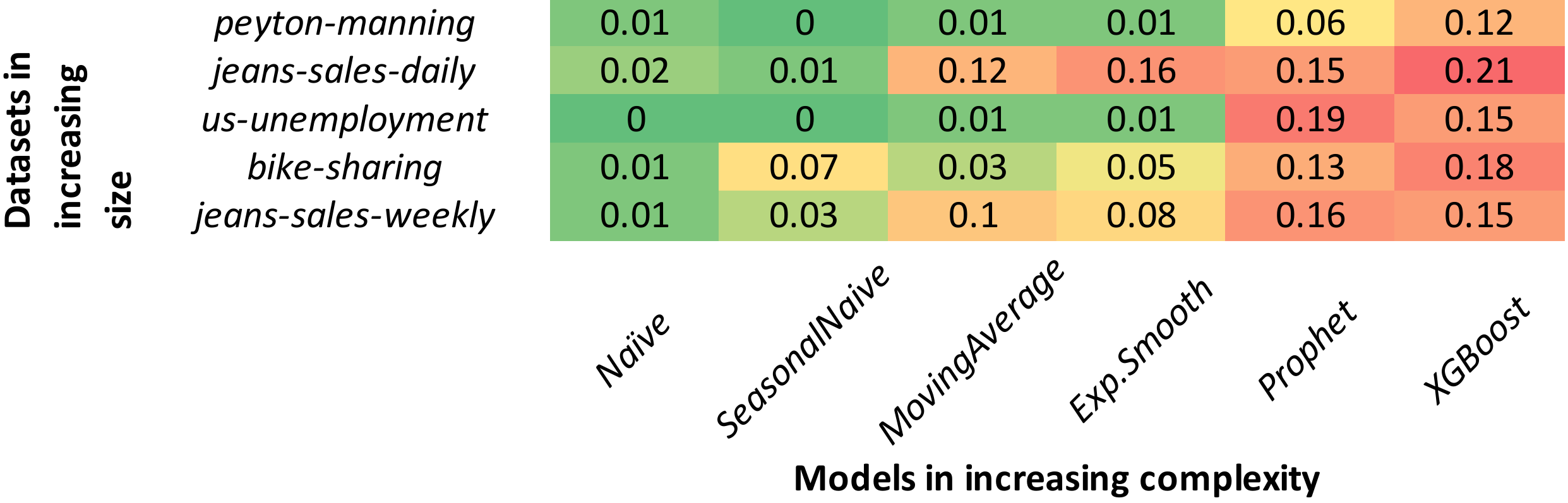}
\caption{MAPE for different models and datasets. In the vertical axis, the dataset size increases from the bottom to the top. In the horizontal axis, the model complexity increases from left to right.}
\label{fig:surrogate_heatmap}
\end{figure}

\begin{figure*}[ht]
     \centering
     \begin{subfigure}[b]{0.33\textwidth}
         \centering
         \includegraphics[width=\textwidth]{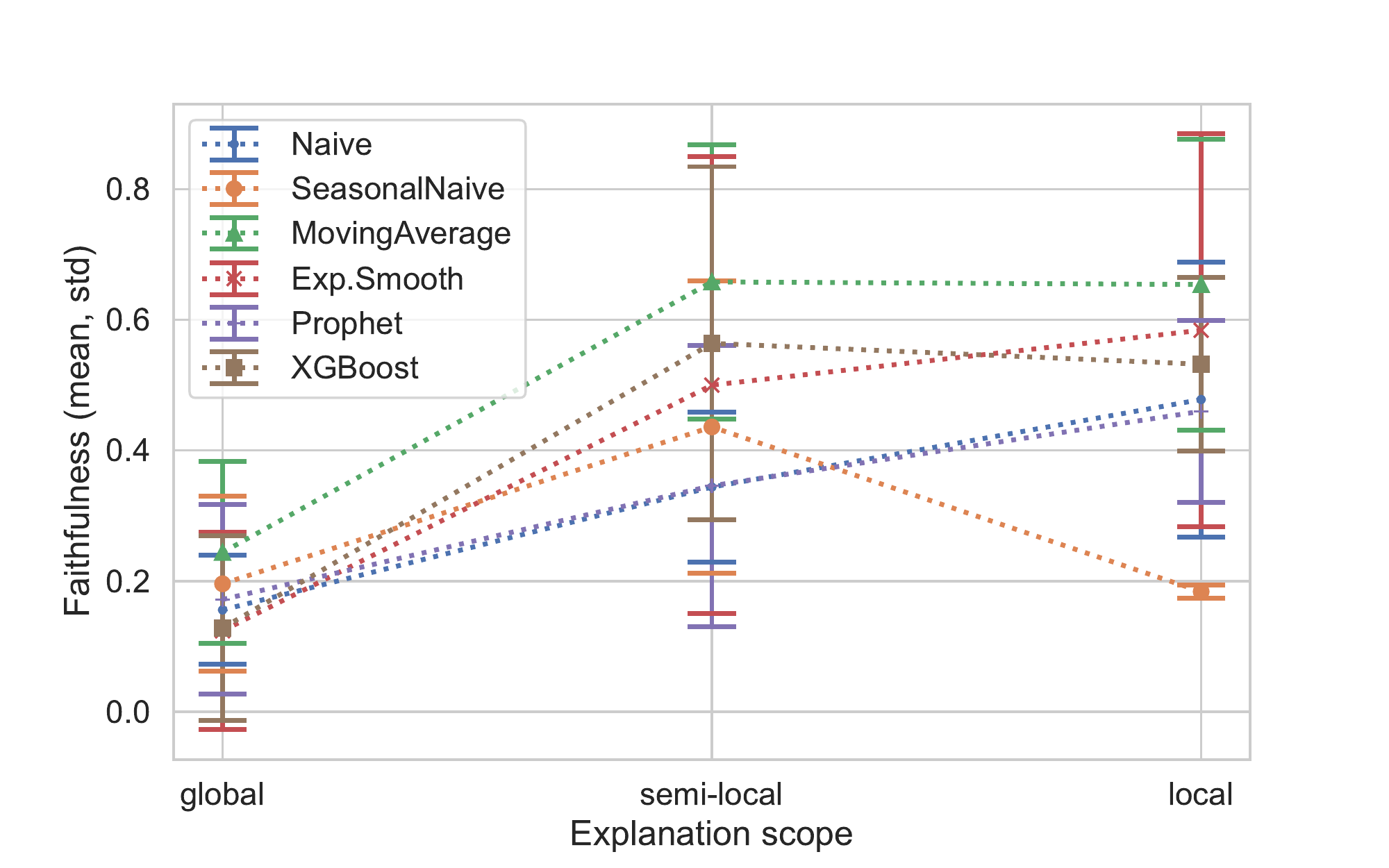}
         \caption{Faithfulness}
         \label{fig:faithfulness_summary}
     \end{subfigure}
     \hfill
     \begin{subfigure}[b]{0.33\textwidth}
         \centering
         \includegraphics[width=\textwidth]{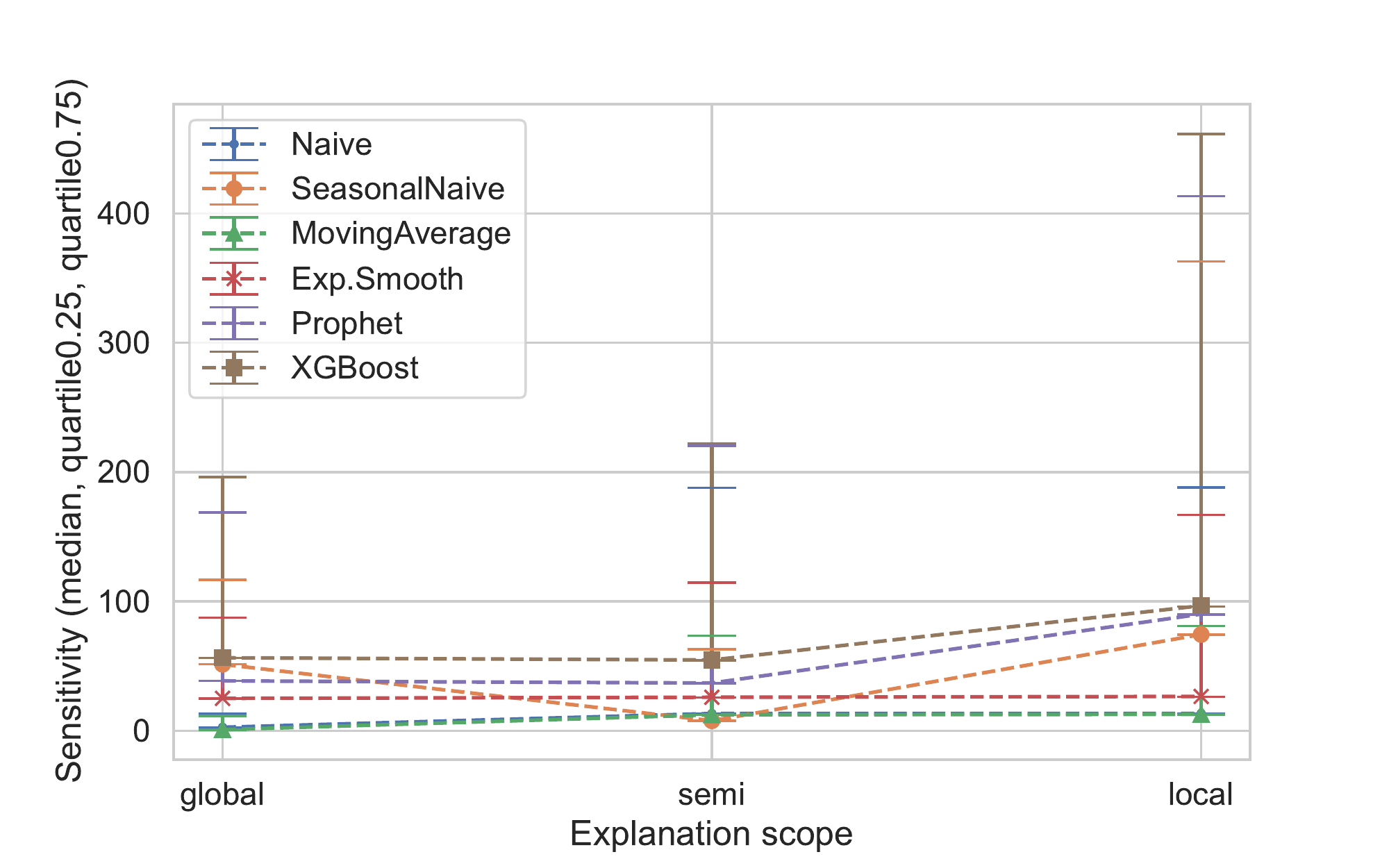}
         \caption{Sensitivity}
         \label{fig:sensitivity_summary}
     \end{subfigure}
     \hfill
     \begin{subfigure}[b]{0.33\textwidth}
         \centering
         \includegraphics[width=\textwidth]{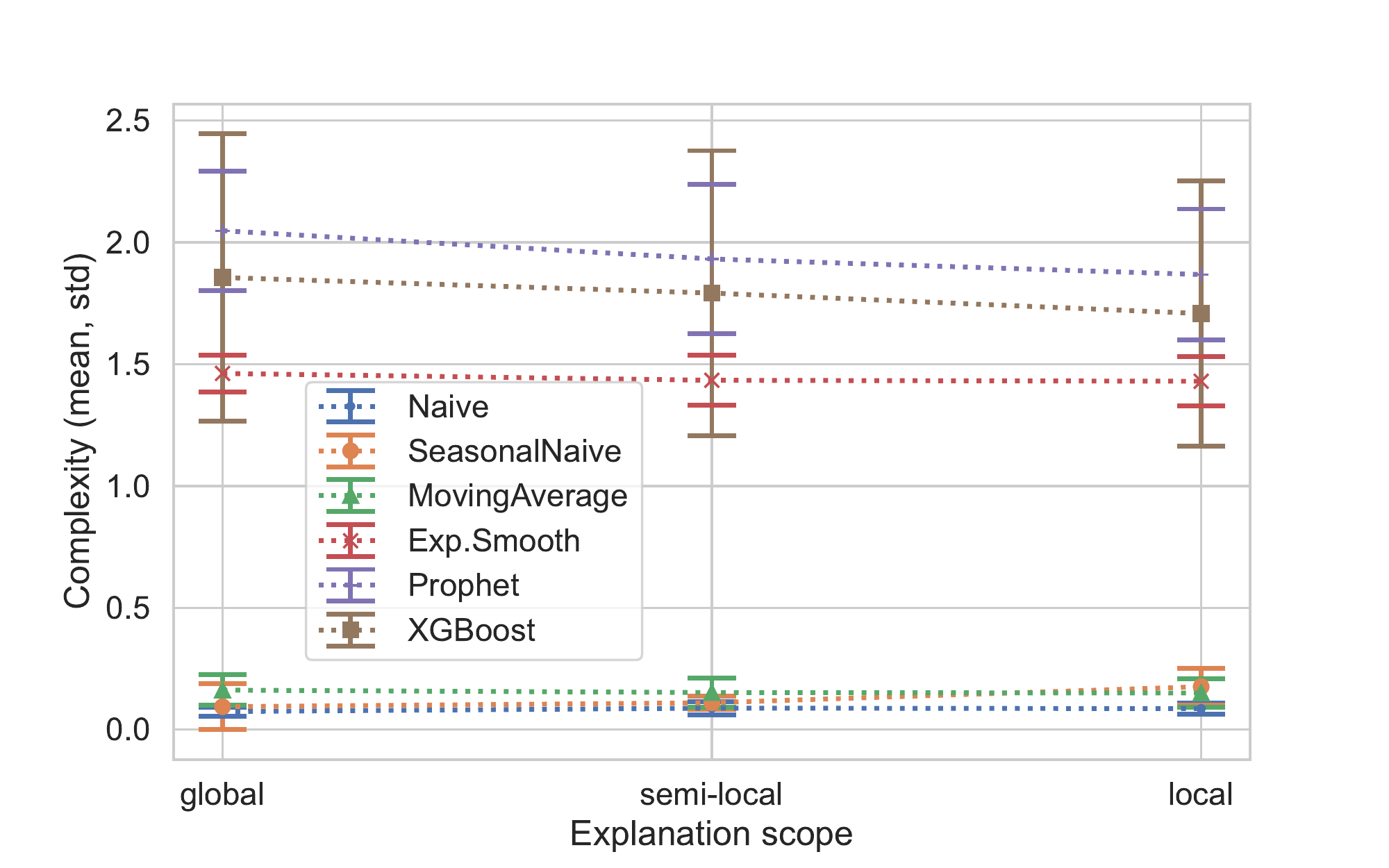}
         \caption{Complexity}
         \label{fig:complexity_summary}
     \end{subfigure}
    \caption{Summary of the evaluation metrics aggregated across all datasets. For faithfulness and complexity, mean is shown with standard deviation as the error bar. For sensitivity, median is shown with 1st and 3rd quartiles as the error bars. Best viewed in color.}
    \label{fig:eval_summary_plots}
\end{figure*}

\begin{figure}[t]
\centering
\begin{subfigure}[b]{\columnwidth}
    \centering
    \includegraphics[width=\columnwidth]{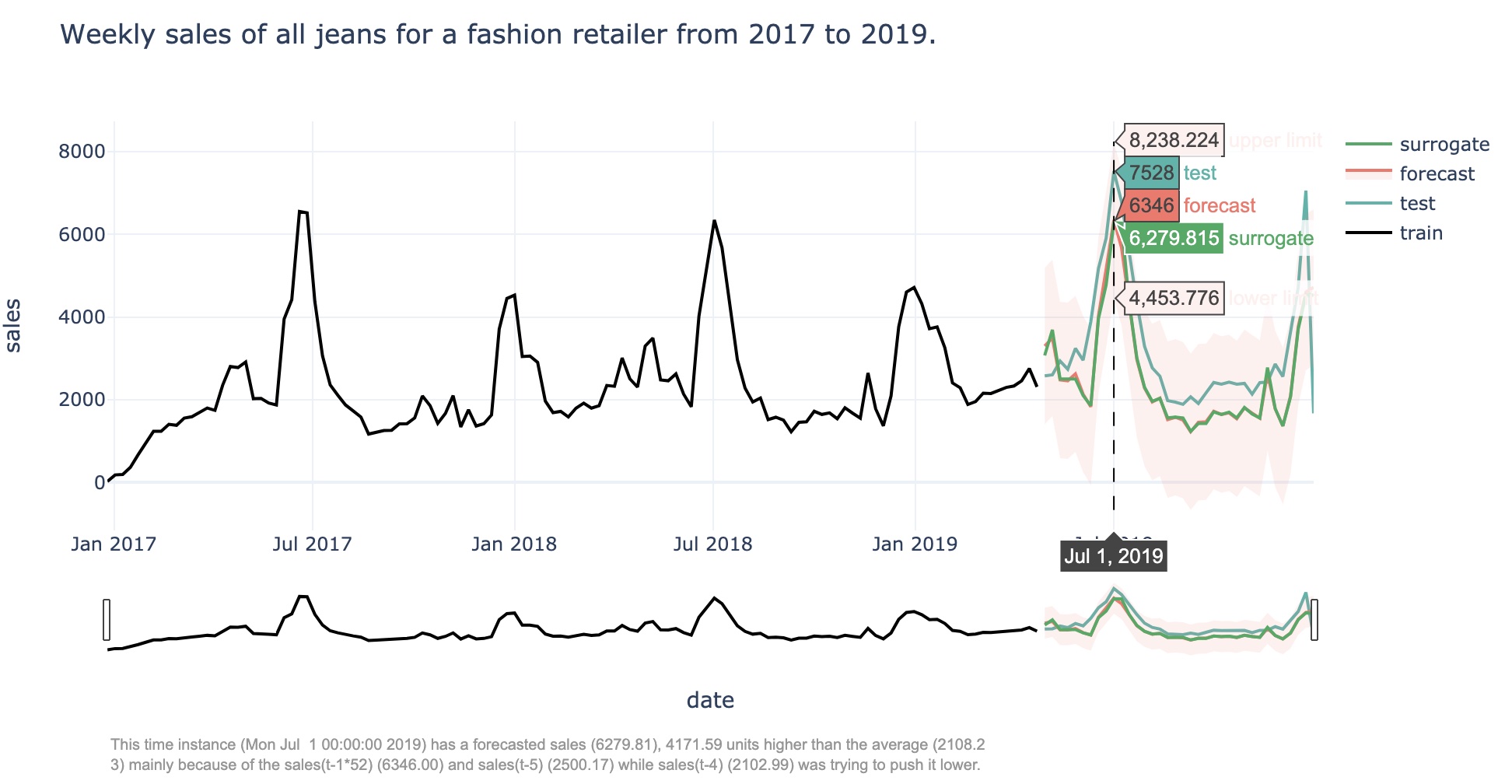}
    \caption{Forecasts from the forecaster and the surrogate model}
\end{subfigure}
\hfill
\begin{subfigure}[b]{\columnwidth}
    \centering
    \includegraphics[width=\columnwidth]{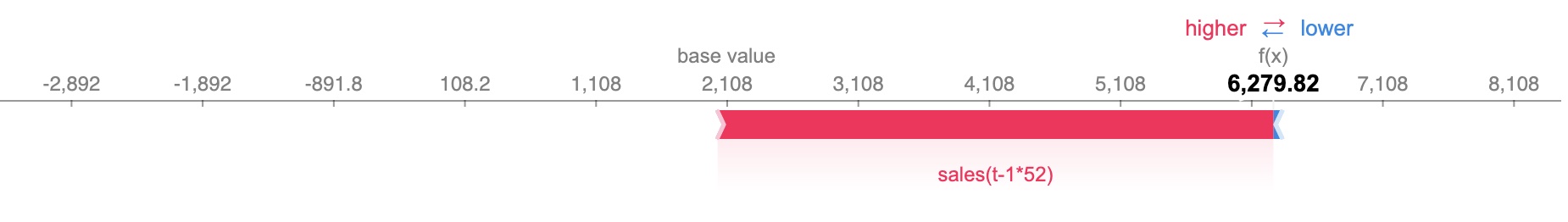}
    \caption{Local explanation}
\end{subfigure}
\hfill
\begin{subfigure}[b]{0.49\columnwidth}
    \centering
    \includegraphics[width=\columnwidth]{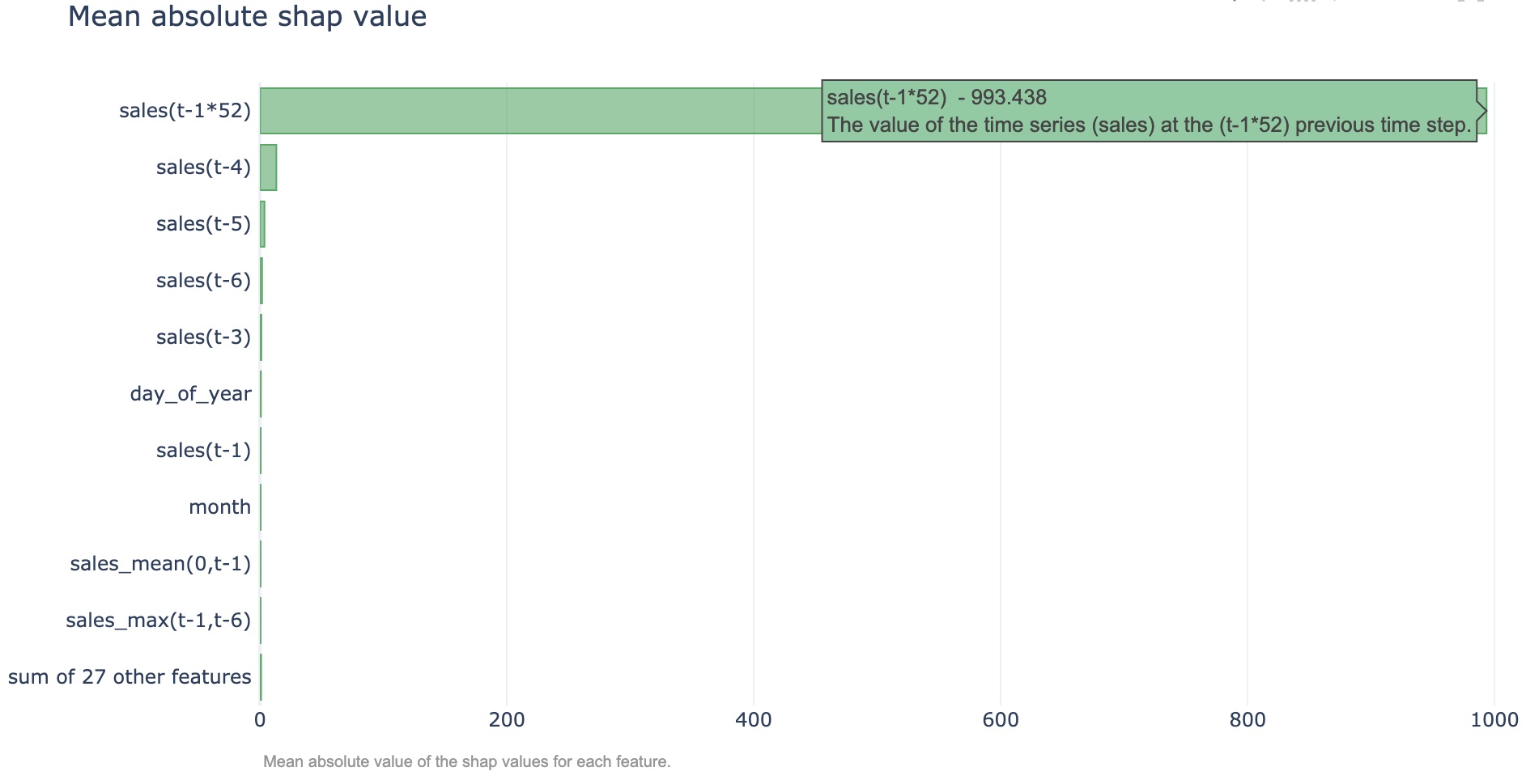}
    \caption{Global explanation}
\end{subfigure}
\hfill
\begin{subfigure}[b]{0.49\columnwidth}
    \centering
    \includegraphics[width=\columnwidth]{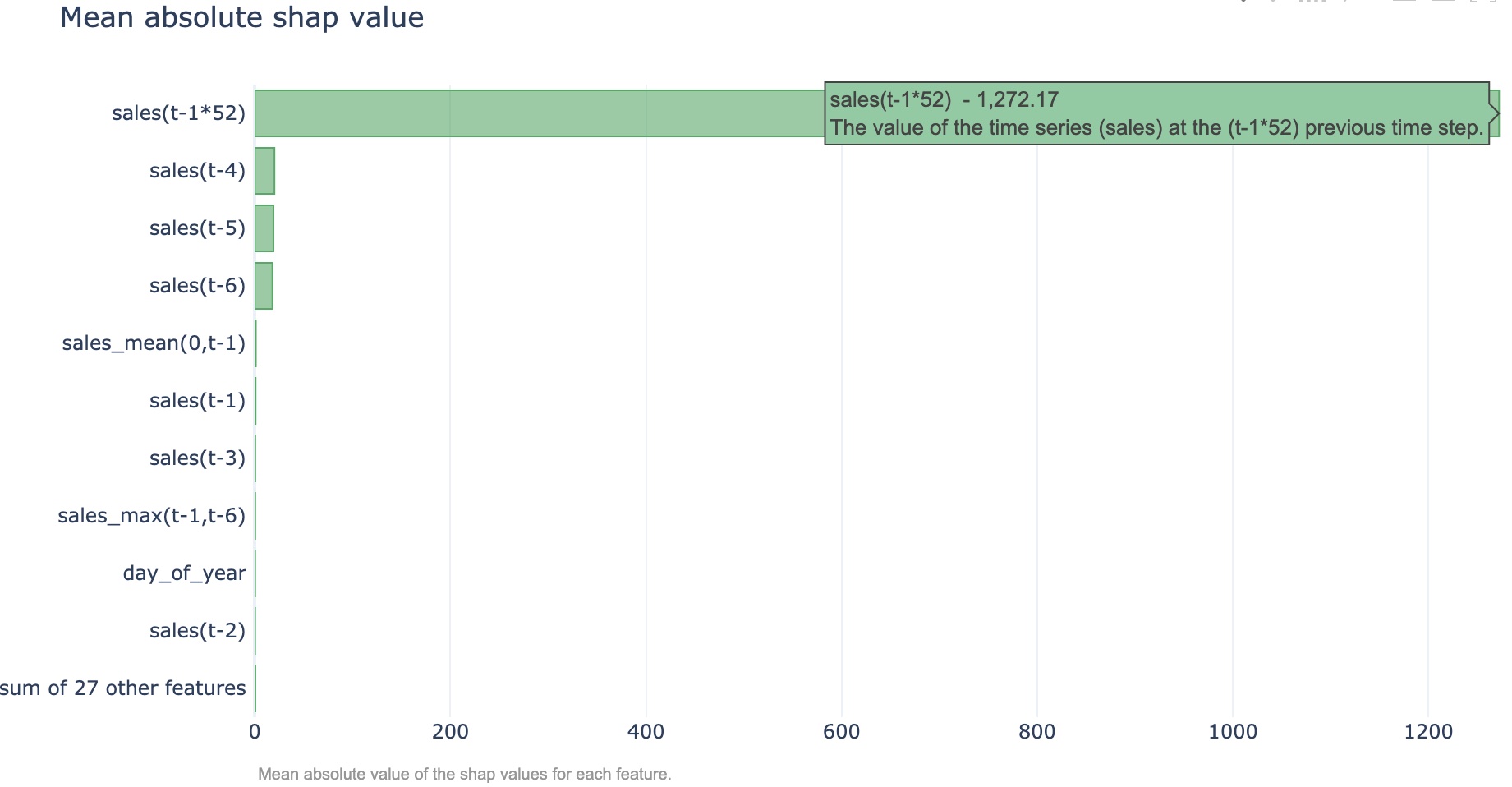}
    \caption{Semi-local explanation}
\end{subfigure} 
\hfill
\begin{subfigure}[b]{0.49\columnwidth}
    \centering
    \includegraphics[width=\columnwidth]{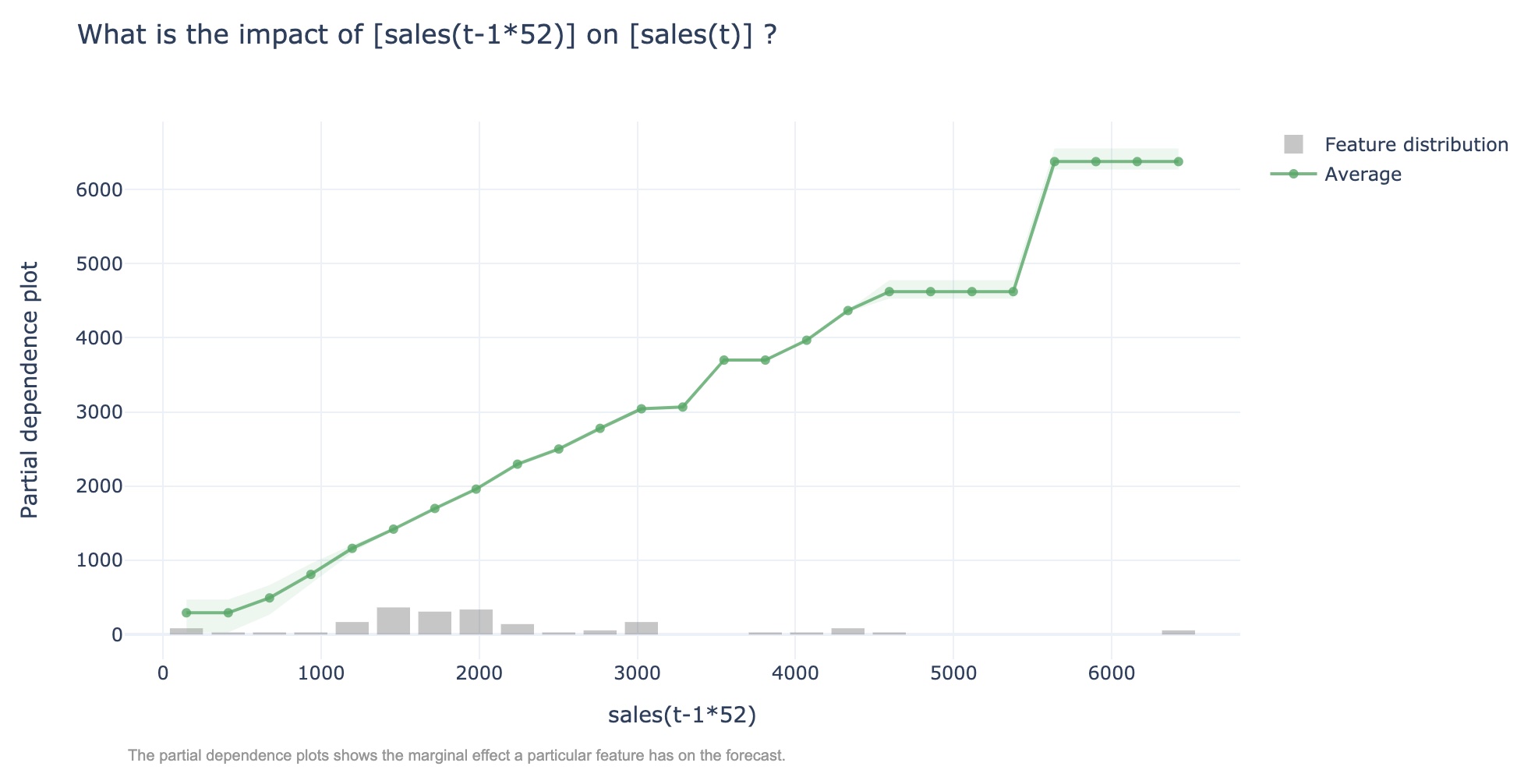}
    \caption{Partial dependence plot}
\end{subfigure}
\hfill
\begin{subfigure}[b]{0.49\columnwidth}
    \centering
    \includegraphics[width=\columnwidth]{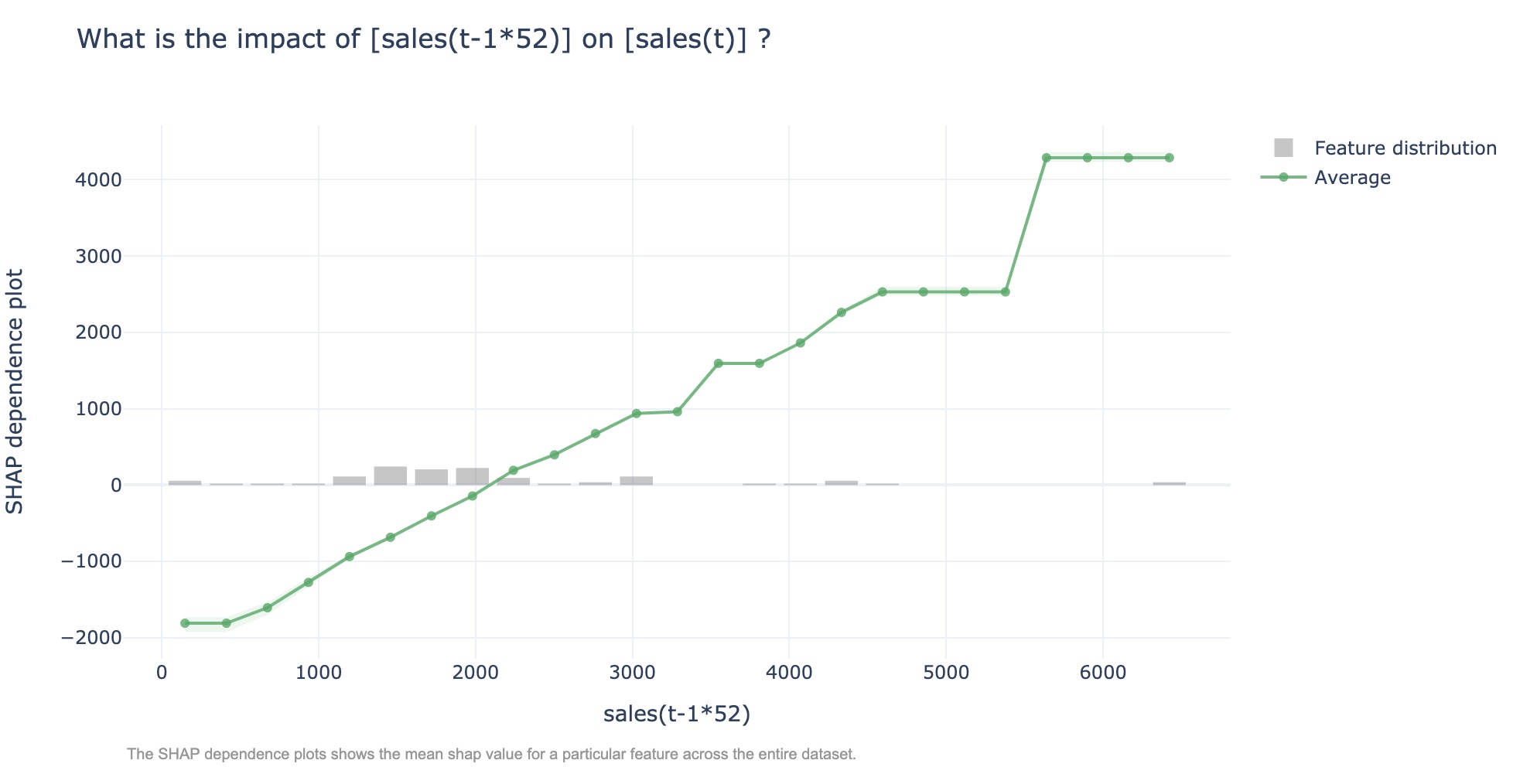}
    \caption{SHAP dependence plot}
 \end{subfigure} 
\caption{Explanations for \textbf{SeasonalNaive(m=52)} forecaster on the \textbf{jeans-sales-weekly} dataset.} 
\label{fig:SeasonalNaive-jeans-sales-weekly}
\end{figure}

\subsection{Accuracy of the surrogate model}

We first validate the accuracy of the surrogate model forecasts from the explainer via backtesting (temporal cross-validation). Using an expanding window approach, we partition the univariate time series into a sequence of train and test datasets. The forecaster and the explainer are trained on the train partition and then evaluated on the test dataset. 
We report various error metrics between the \textbf{forecasts from the original forecaster} and the \textbf{forecasts made by the surrogate model of the explainer}, viz. MAE (Mean Absolute Error), RMSE (Root Mean Squared Error), MAPE (Mean Absolute Percentage Error), and MASE (Mean Absolute Scaled Error)~\cite{mase_paper}. Table~\ref{tab:metrics-surrogate} (in \S~\ref{sec:Reproducibility} Reproducibility) shows the accuracy scores between the forecaster and the surrogate model for the six forecasters and five datasets.

Figure~\ref{fig:surrogate_heatmap} shows the variation in MAPE with increasing size of the datasets and increasing complexity of the models.
It is evident that the MAPE increases with an increase in the model complexity.
This might be because it is harder for the surrogate model to mimic a complex forecaster (e.g. XGBoost) than a simpler one (e.g., Naive).
However, there is no such pattern in the variation of MAPE with increase in the dataset size. 
This is particularly beneficial since it conveys that the dataset size does not affect the surrogate model training.

\subsection{Evaluating explanations}
We evaluate the feature-based explanations in terms of the (high) faithfulness, (low) sensitivity and (low) complexity metrics defined in \S~\ref{sec:metrics}. In the Reproducibility section, \S~\ref{sec:Reproducibility}, we provide various metrics for global, semi-local and local explanations for all the forecasters and the datasets in Table~\ref{tab:metrics-global}, \ref{tab:metrics-semilocal} and \ref{tab:metrics-local}. The metrics for the local explanations were averaged over the entire forecast horizon.

Figure~\ref{fig:eval_summary_plots} provides a graphical summary of Table~\ref{tab:metrics-global}, \ref{tab:metrics-semilocal} and \ref{tab:metrics-local} by aggregating the metric values across all the datasets\footnote{For sensitivity, we do not show mean (std) because its value depends on the absolute value of the data (see \S~\ref{sec:metrics}). Instead, we show median with 1st and 3rd quartiles as the error bars.}.
We can observe that, for all the forecasters, the faithfulness of semi-local and local explanations are higher than that of global explanations.
A closer look revelas that the faithfulness of a semi-local explanation is always higher than the faithfulness of a global explanation.
However, when we move to local explanations (from semi-local), there is no definite pattern in change in the faithfulness metric.
The complexity of the explanations do not vary much with the scope of explanation, as can be seen in Figure~\ref{fig:complexity_summary}.
This might refer to an implicit relationship between the complexity of the forecast explanation and that of the forecaster.
The median sensitivity curves stay (approximately) flat for all the forecasters.

Moreover, we can notice that a particular forecaster might not achieve the best performance in all the metrics, and this highlights the complementary nature of the metrics defined in \S~\ref{sec:metrics}.
For example, in Table~\ref{tab:metrics-global}, for ``jeans-sales-weekly'' dataset, MovingAverage forecaster achieves the highest faithfulness and the lowest sensitivity, but the lowest complexity is obtained by the Naive forecaster.
This type of behaviors is also observed in local and semi-local explanations (Table~\ref{tab:metrics-local} and Table~\ref{tab:metrics-semilocal}).

A closer look on Figure~\ref{fig:eval_summary_plots} (and  on Table~\ref{tab:metrics-global}, \ref{tab:metrics-local} and \ref{tab:metrics-semilocal}) reveals that the MovingAverage forecaster achieves the average highest faithfulness (0.244 for global, 0.654 for local, and 0.658 for semi-local explanations) across all datasets.
The Naive forecaster obtains the average lowest complexity (0.074 for global, 0.086 for local, and 0.088 for semi-local explanations) across all datasets. 
This might be because the inherent simplicity of the Naive forecaster results in the fractional distribution of the feature importance scores to have low entropy (see \S~\ref{sec:metrics}).

However, the median lowest sensitivity is shared among MovingAverage and SeasonalNaive forecasters across different scopes of explanations.
The MovingAverage forecaster has the lowest median sensitivity for global (0.8) and local (12.64) explanations, but the SeasonalNaive attains the lowest value (7.7) for semi-local explanation.

Another noticeable observation from our experiments is that the complex (and possibly non-linear) models (like Prophet and XGBoost) have high complexity, which is expected. 
However, they might achieve better faithfulness and sensitivity than a simpler model. 
For example, in Table~\ref{tab:metrics-global}, Prophet attains higher faithfulness than a Naive forecaster, and in Table~\ref{tab:metrics-semilocal}, XGBoost achieves lower sensitivity and higher faithfulness than a Naive forecaster in multiple datasets.

\begin{figure}[t]
\centering
\begin{subfigure}[b]{\columnwidth}
    \centering
    \includegraphics[width=\columnwidth]{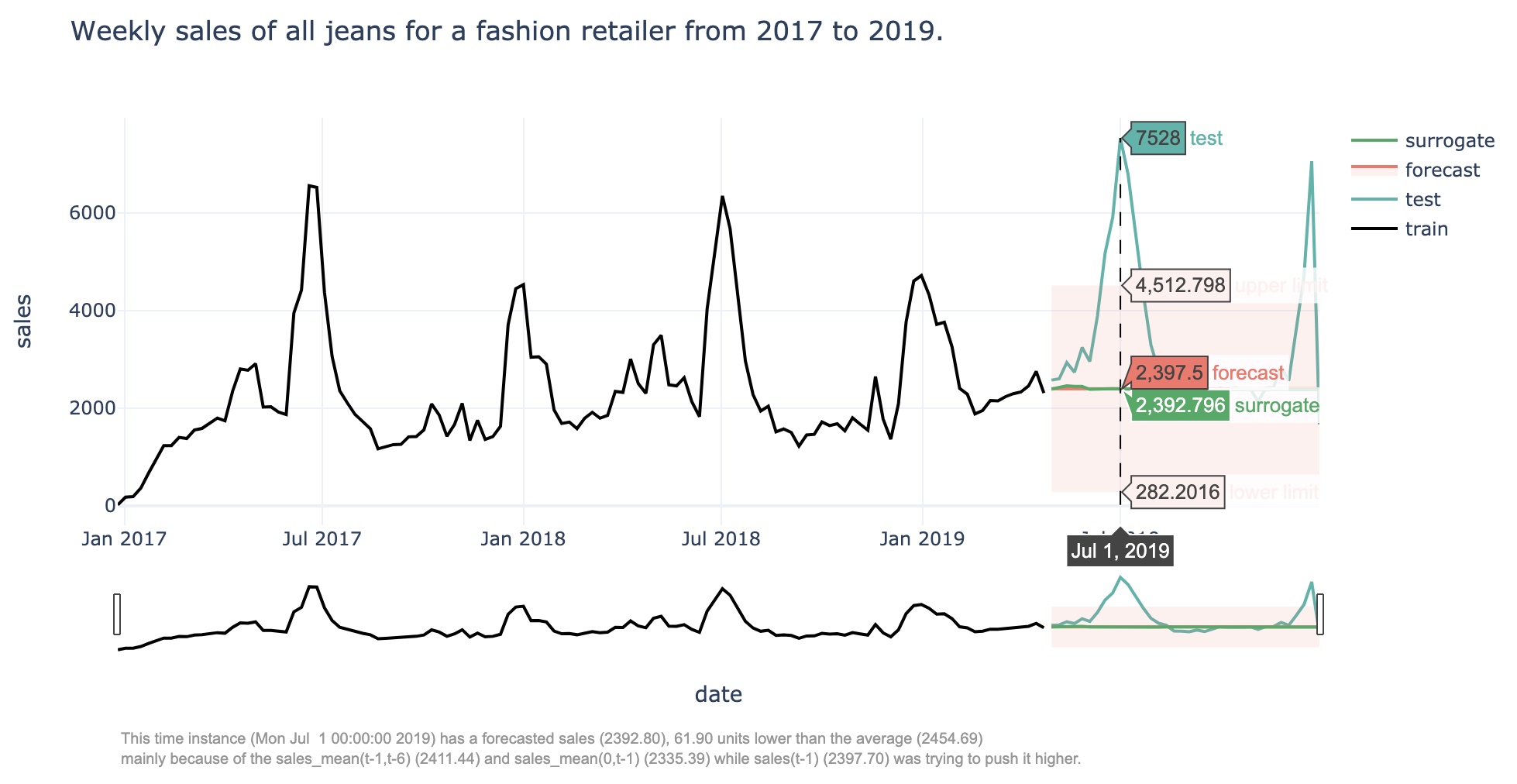}
    \caption{Forecasts from the forecaster and the surrogate model}
\end{subfigure}
\hfill
\begin{subfigure}[b]{\columnwidth}
    \centering
    \includegraphics[width=\columnwidth]{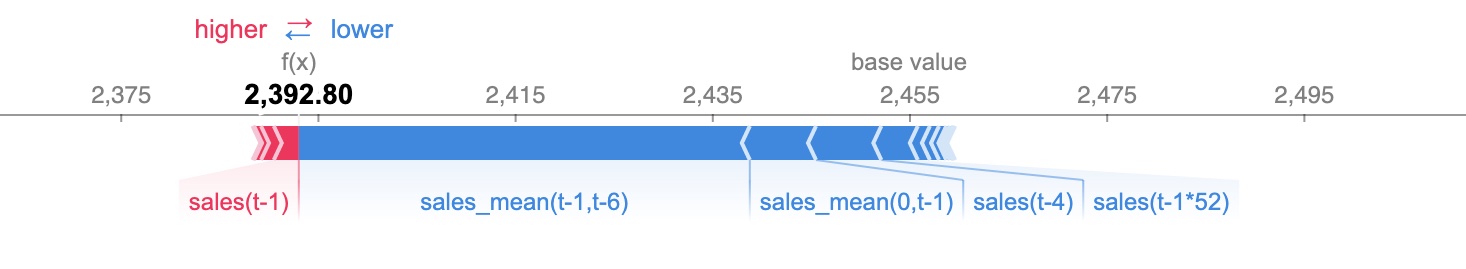}
    \caption{Local explanation}
\end{subfigure}
\hfill
\begin{subfigure}[b]{0.49\columnwidth}
    \centering
    \includegraphics[width=\columnwidth]{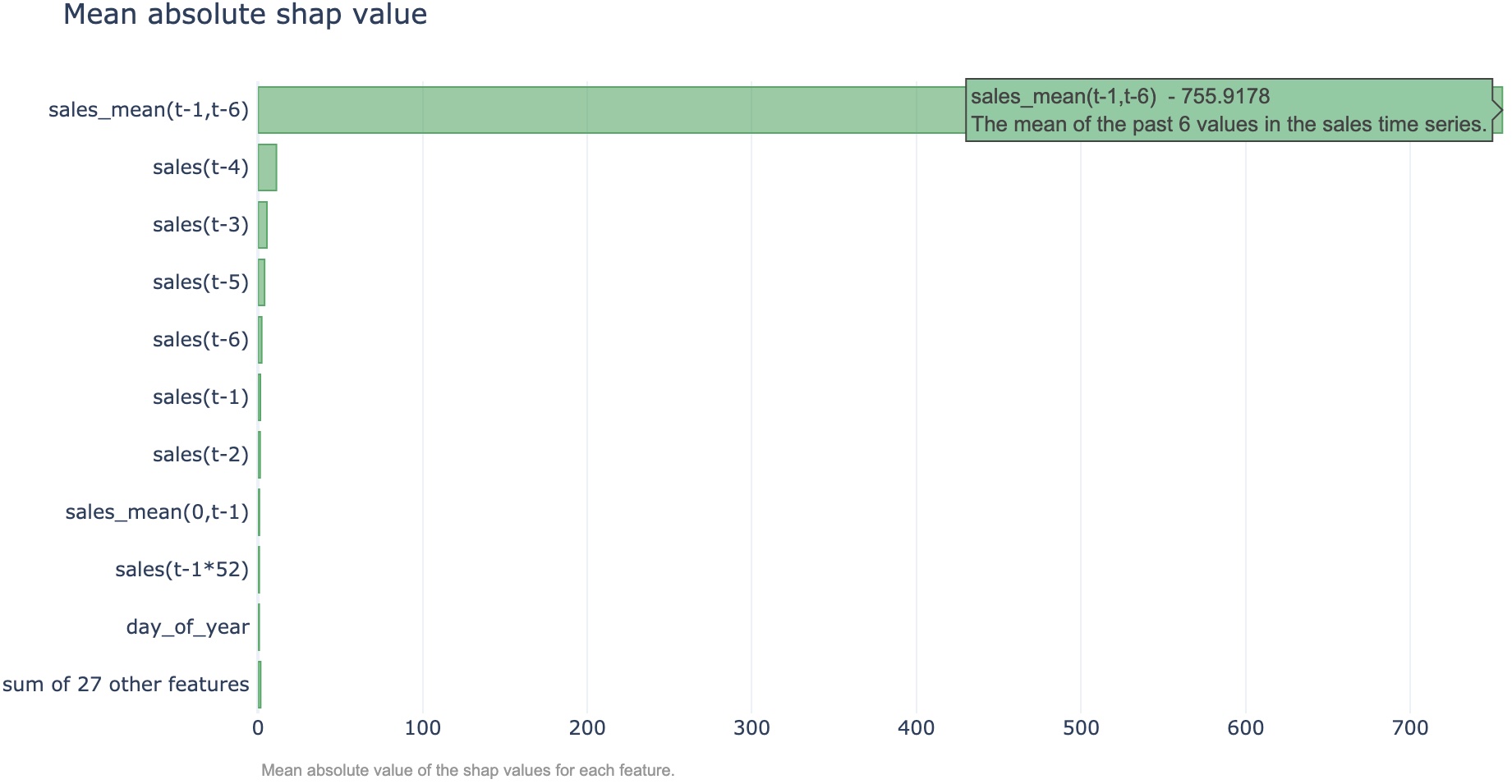}
    \caption{Global explanation}
\end{subfigure}
\hfill
\begin{subfigure}[b]{0.49\columnwidth}
    \centering
    \includegraphics[width=\columnwidth]{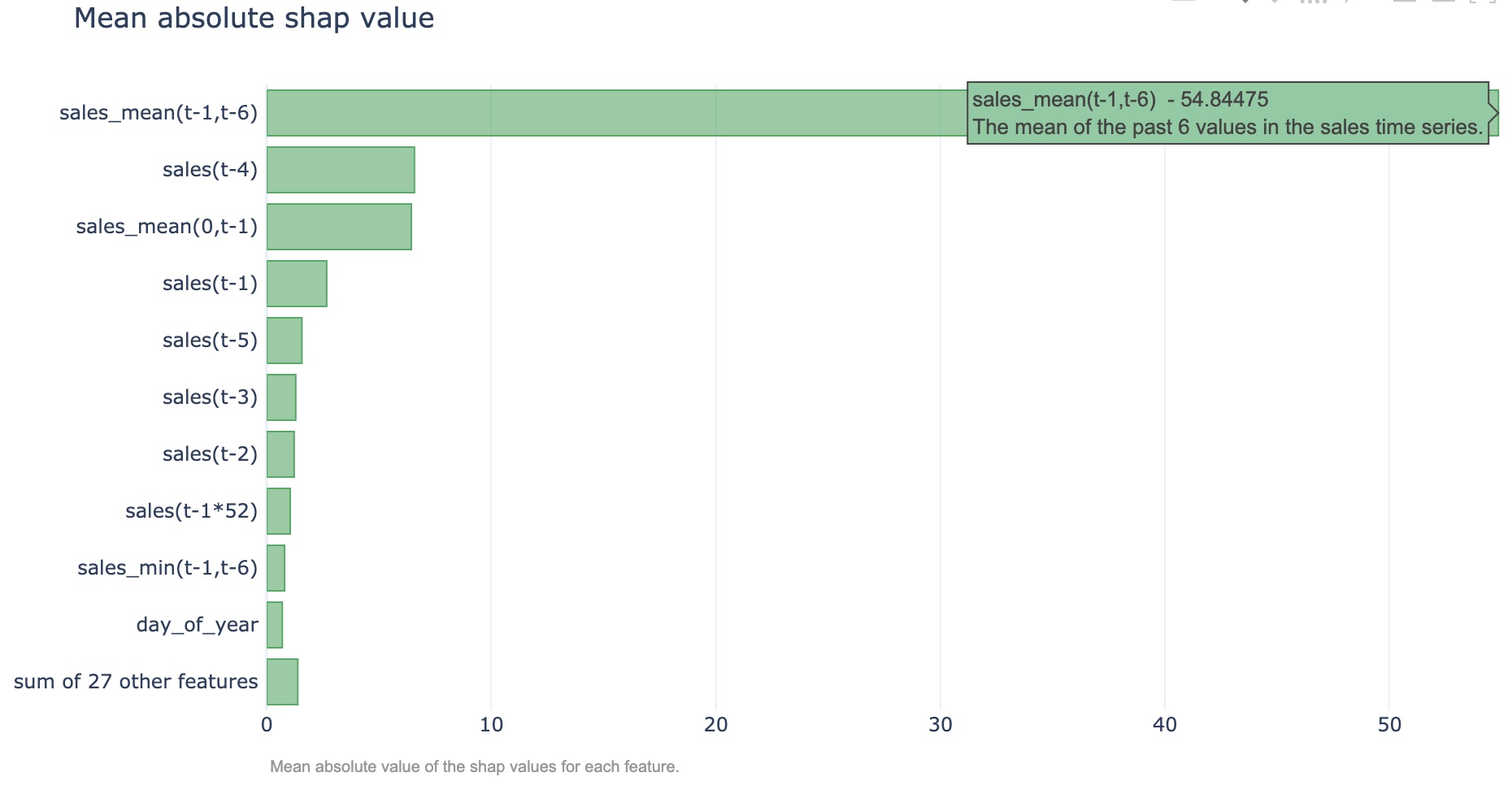}
    \caption{Semi-local explanation}
\end{subfigure} 
\hfill
\begin{subfigure}[b]{0.49\columnwidth}
    \centering
    \includegraphics[width=\columnwidth]{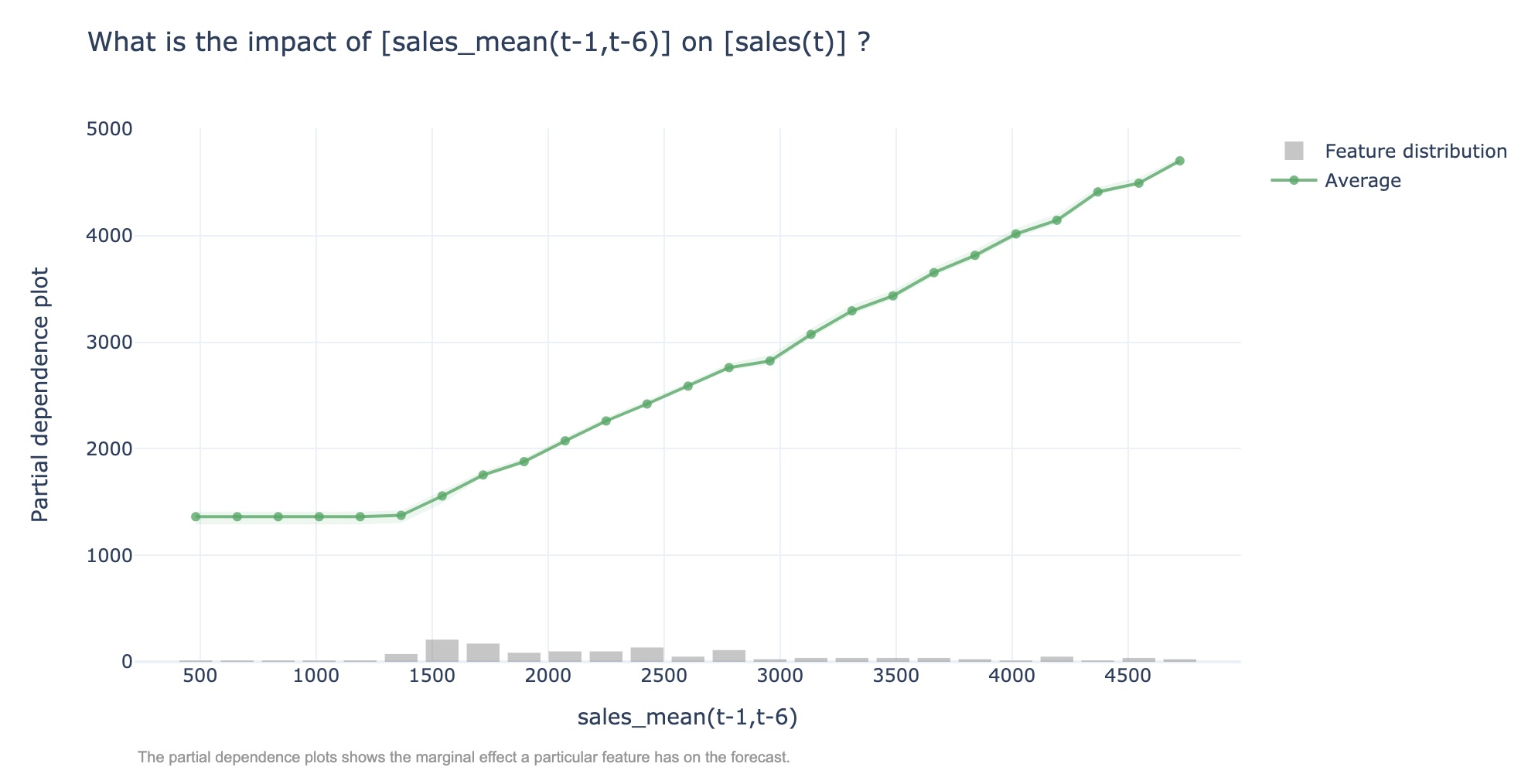}
    \caption{Partial dependence plot}
\end{subfigure}
\hfill
\begin{subfigure}[b]{0.49\columnwidth}
    \centering
    \includegraphics[width=\columnwidth]{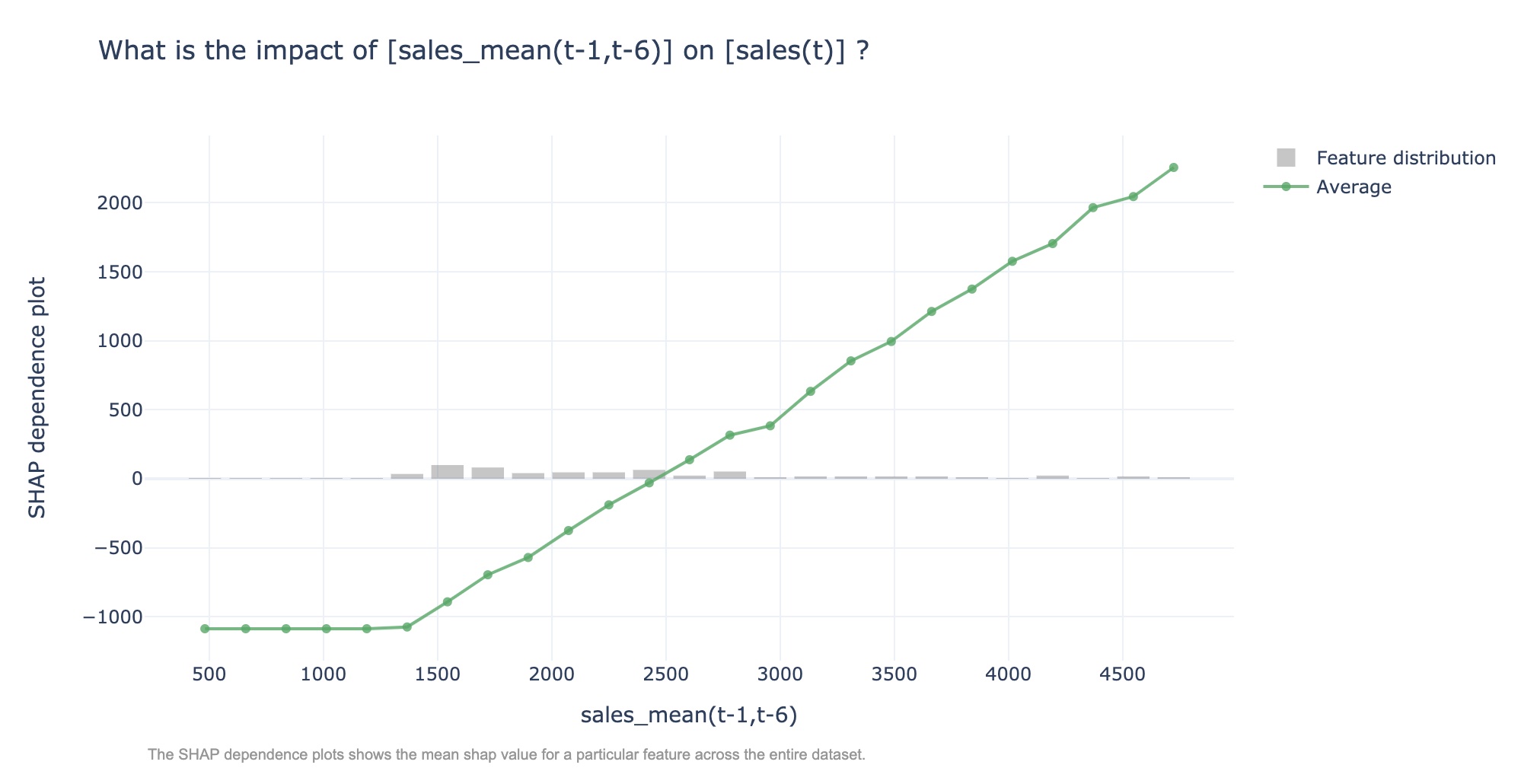}
    \caption{SHAP dependence plot}
 \end{subfigure} 
\caption{Explanations for \textbf{MovingAverage(k=6)} forecaster on the \textbf{jeans-sales-weekly} dataset.} 
\label{fig:MovingAverage-jeans-sales-weekly}
\end{figure}

\begin{figure}[ht]
\centering
\begin{subfigure}[b]{\columnwidth}
    \centering
    \includegraphics[width=\columnwidth]{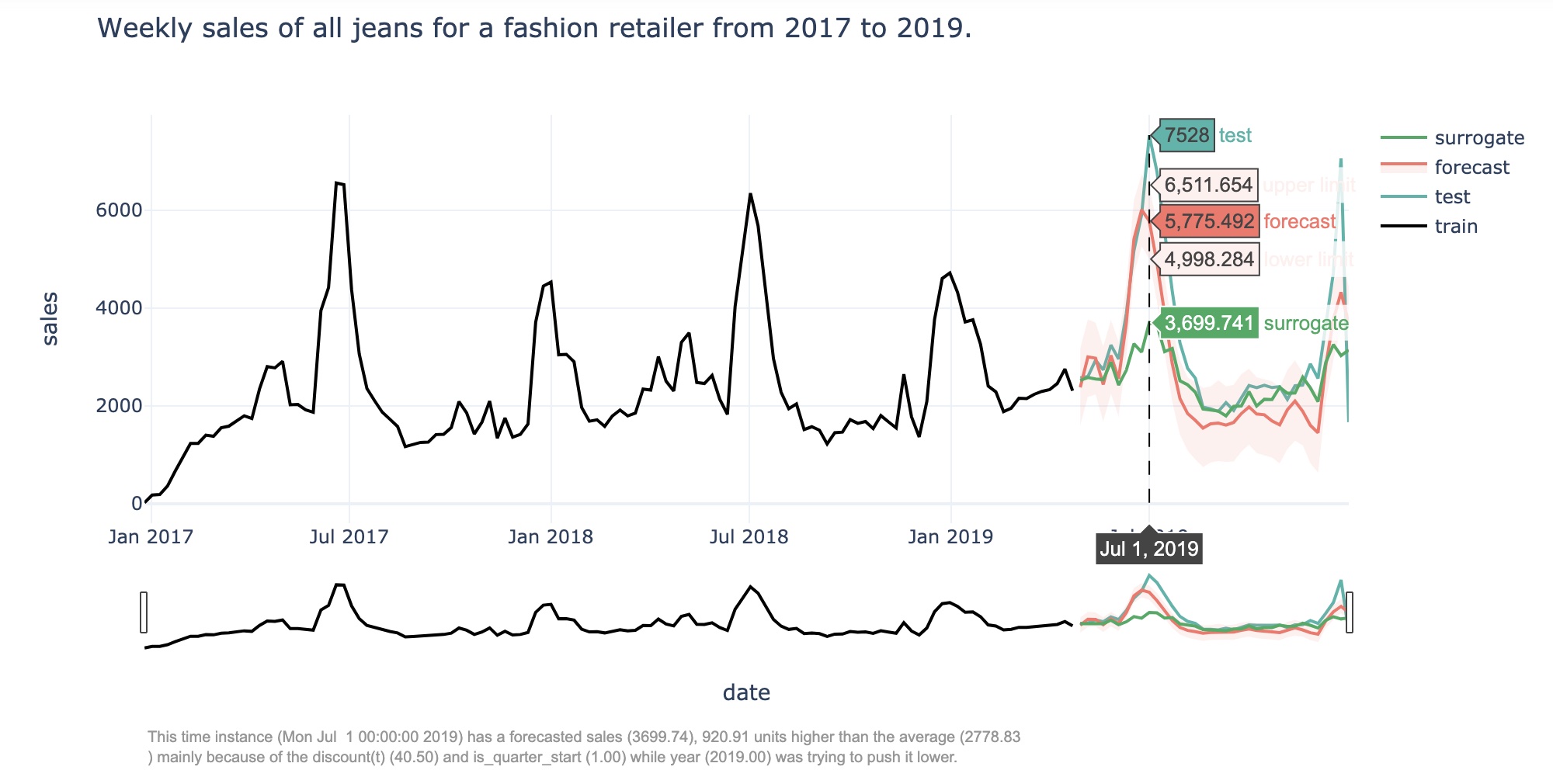}
    \caption{Forecasts from the forecaster and the surrogate model}
\end{subfigure}
\hfill
\begin{subfigure}[b]{\columnwidth}
    \centering
    \includegraphics[width=\columnwidth]{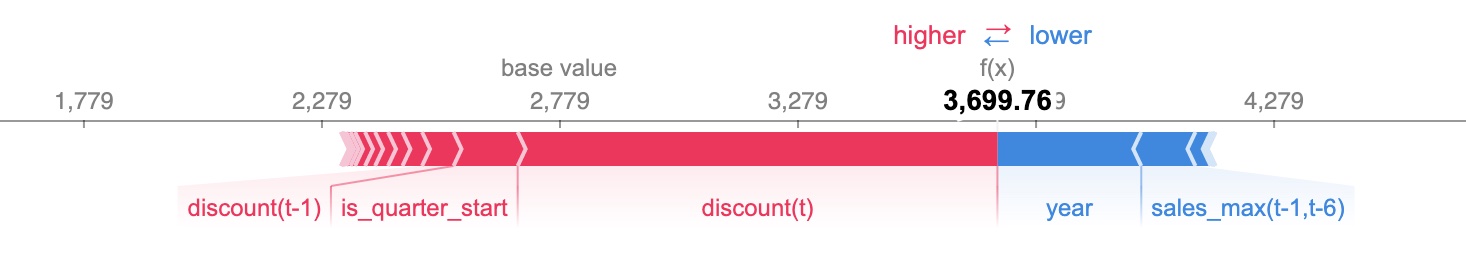}
    \caption{Local explanation}
\end{subfigure}
\hfill
\begin{subfigure}[b]{0.49\columnwidth}
    \centering
    \includegraphics[width=\columnwidth]{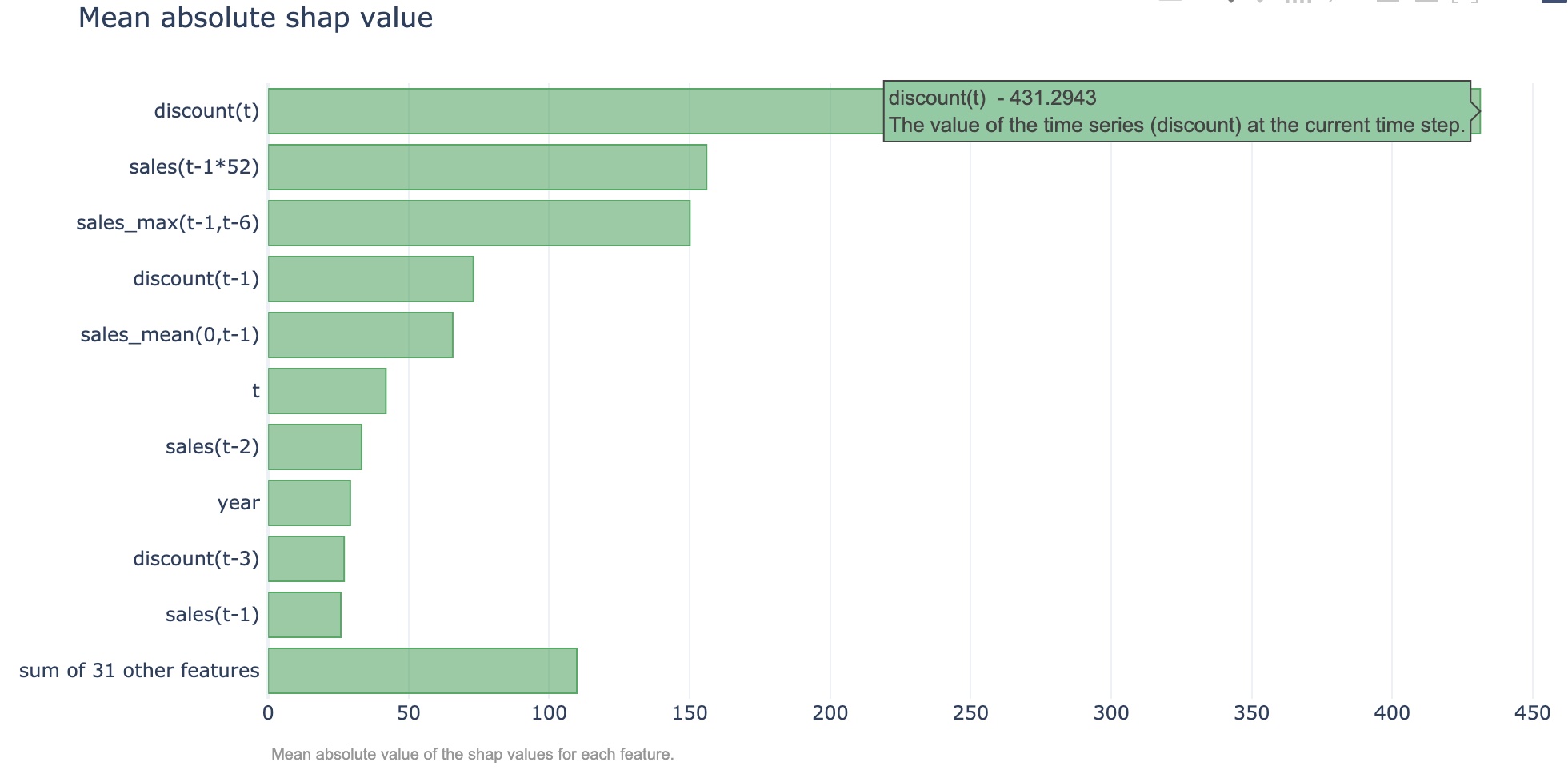}
    \caption{Global explanation}
\end{subfigure}
\hfill
\begin{subfigure}[b]{0.49\columnwidth}
    \centering
    \includegraphics[width=\columnwidth]{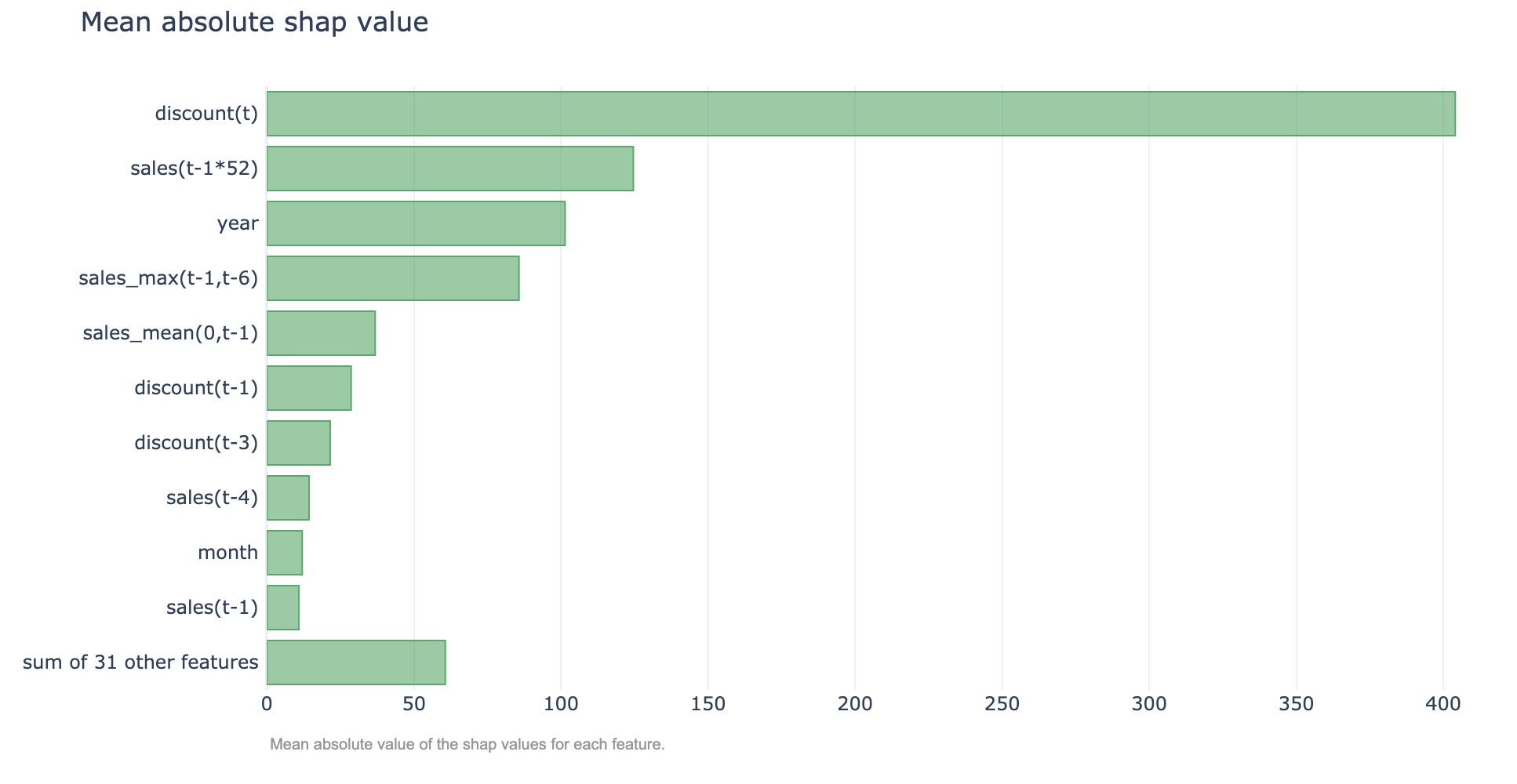}
    \caption{Semi-local explanation}
\end{subfigure} 
\hfill
\begin{subfigure}[b]{0.49\columnwidth}
    \centering
    \includegraphics[width=\columnwidth]{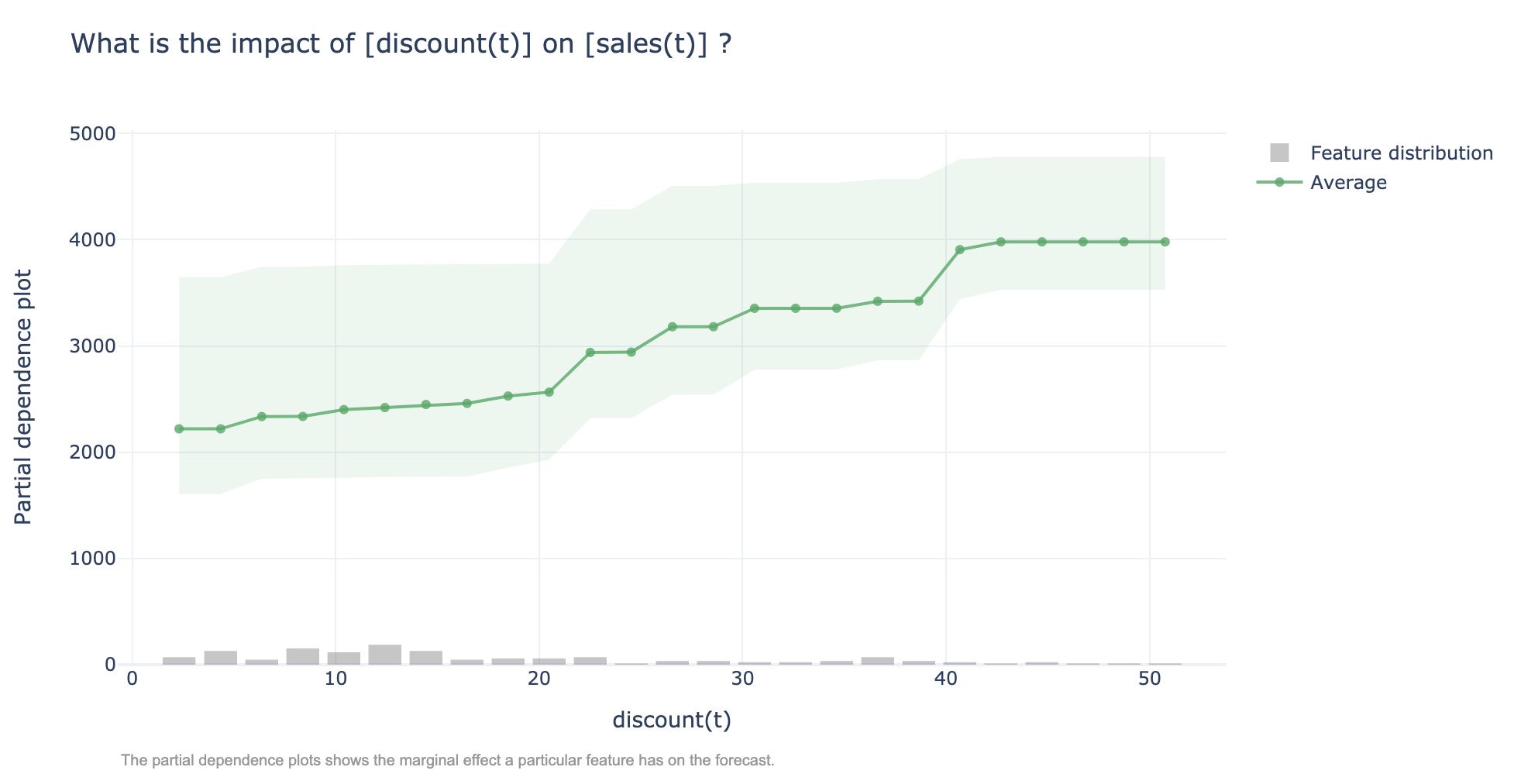}
    \caption{Partial dependence plot}
\end{subfigure}
\hfill
\begin{subfigure}[b]{0.49\columnwidth}
    \centering
    \includegraphics[width=\columnwidth]{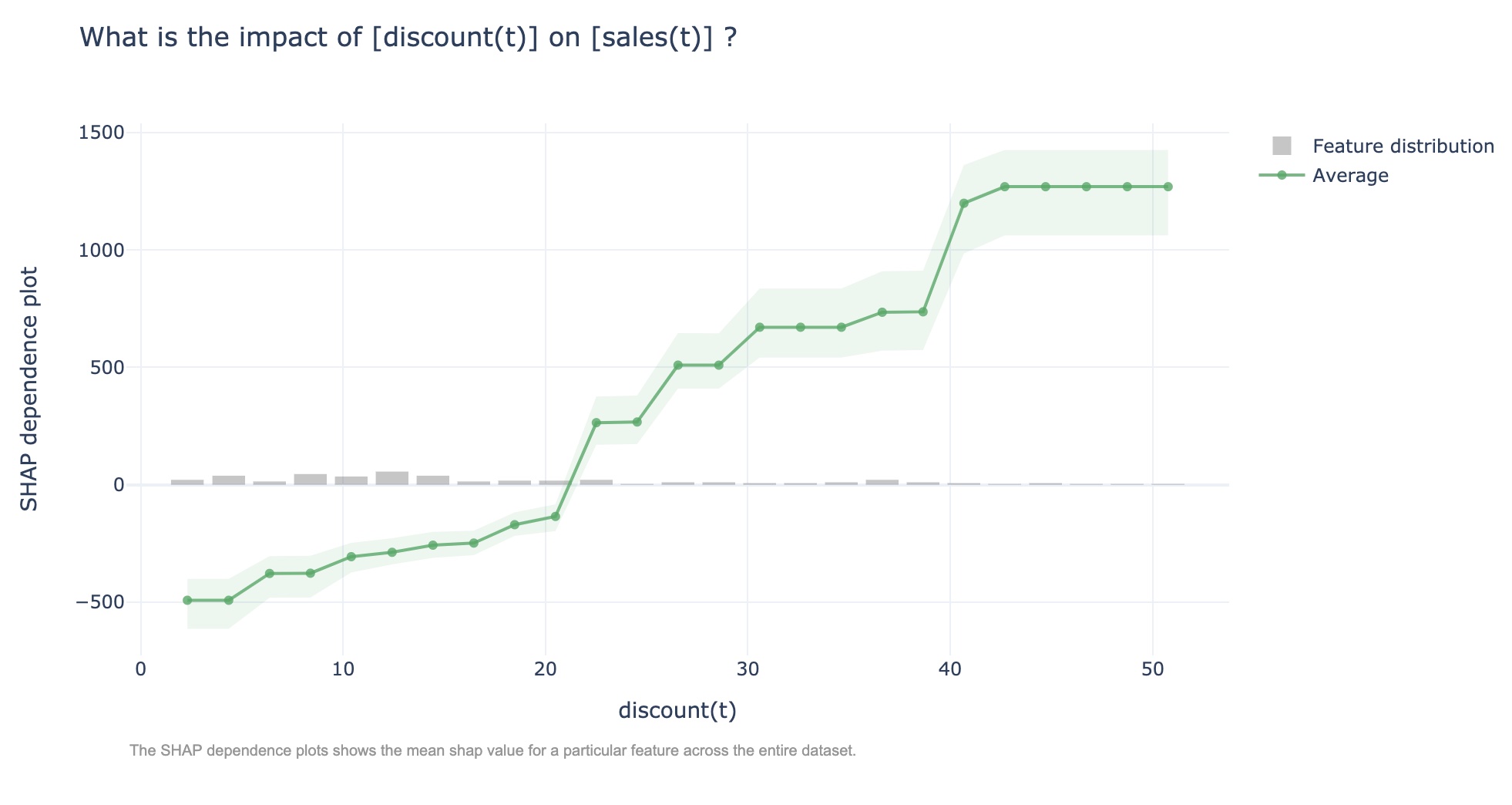}
    \caption{SHAP dependence plot}
 \end{subfigure} 
\caption{Explanations for \textbf{Prophet} forecaster on the \textbf{jeans-sales-weekly} dataset.} 
\label{fig:Prophet-jeans-sales-weekly}
\end{figure}

\subsection{TsSHAP Illustrations}
Figure~\ref{fig:SeasonalNaive-jeans-sales-weekly}, \ref{fig:MovingAverage-jeans-sales-weekly}, and \ref{fig:Prophet-jeans-sales-weekly} illustrate a few local, semi-local, and global explanations for the jeans-sales-daily dataset for all different forecasters. 
Note that the PDP and SDP plots are only shown for a local scope.
It can be seen that the explanations reflect the underlying aspects the forecaster is able to learn from the data.
From the local, semi-local, and global explanations of Figure~\ref{fig:SeasonalNaive-jeans-sales-weekly}, we can see that the main contributing factor for the forecast is the seasonal history at $(t-52)$ weeks time, which is expected from a seasonal naive forecaster. The partial dependence plot also depicts a linear relationship between the \texttt{sales(t-52)} and forecaster output.
From the local explanation of Figure~\ref{fig:MovingAverage-jeans-sales-weekly}, we can see that the mean value of the past 6 observations is primarily responsible for bringing the forecast down. Other explanations depict the same story.
From the local explanation of Figure~\ref{fig:Prophet-jeans-sales-weekly}, we can see that, mainly, \texttt{discount(t)} and \texttt{is\_quarters\_start} are pushing the forecast up, but \texttt{year} and \texttt{sales\_max(t-1, t-6)} are pushing the forecast down. The Prophet forecaster supports external regressors, and here, it is picking discount as the primary explanatory factor. The global explanation clearly shows the main contributing features responsible for the forecast across the entire time series dataset.
Some other illustrations are provided in \S~\ref{sec:Reproducibility}.

\section{Conclusions and future work}
We proposed a robust interpretable feature-based explainability algorithm for time-series forecasting. The method is model agnostic and does not require access to the model internals. Moreover, we formalized the notion of local, semi-local, and global explanations in the context of time series forecasting. We validated the correctness of the explanations by applying TsSHAP on multiple forecasters. 
We demonstrated the robustness of TsSHAP by evaluating the explanations with faithfulness and sensitivity metrics that work on multiple perturbed versions of a time series. 
Following are the main observations from the experiments.
\begin{itemize}
\item The TsSHAP explanations for the inherently interpretable forecasters (like Naive, Seasonal Naive, Moving Average, and Exponential Smoothing) were found to be correct, \ie they follow the modeling strategy of the forecaster.
\item The surrogate model prediction accuracy decreases with increasing model complexity.
\item The TsSHAP explanations are \textit{more faithful} in the local and semi-local scopes.
\item The \textit{complexity} of an explanation is stable over different explanation scopes.
\item Although TsSHAP tends to generate complex explanations for a complex model, the explanations might have better faithfulness and sensitivity scores than a simpler model. 
\end{itemize}

TsSHAP can also be applied to derive interpretable insights of complex neural network-based models. TsSHAP can be extended to explain the prediction interval. Instead of regressing on the mean forecast from the forecaster, we regress on the prediction interval. We also plan to extend the above algorithm to multivariate time series forecasting models.

\bibliographystyle{ACM-Reference-Format}
\bibliography{TsSHAP}


\begin{thebibliography}{31}


\ifx \showCODEN    \undefined \def \showCODEN     #1{\unskip}     \fi
\ifx \showDOI      \undefined \def \showDOI       #1{#1}\fi
\ifx \showISBNx    \undefined \def \showISBNx     #1{\unskip}     \fi
\ifx \showISBNxiii \undefined \def \showISBNxiii  #1{\unskip}     \fi
\ifx \showISSN     \undefined \def \showISSN      #1{\unskip}     \fi
\ifx \showLCCN     \undefined \def \showLCCN      #1{\unskip}     \fi
\ifx \shownote     \undefined \def \shownote      #1{#1}          \fi
\ifx \showarticletitle \undefined \def \showarticletitle #1{#1}   \fi
\ifx \showURL      \undefined \def \showURL       {\relax}        \fi
\providecommand\bibfield[2]{#2}
\providecommand\bibinfo[2]{#2}
\providecommand\natexlab[1]{#1}
\providecommand\showeprint[2][]{arXiv:#2}

\bibitem[\protect\citeauthoryear{Assaf and Schumann}{Assaf and
  Schumann}{2019}]%
        {assaf2019explainable}
\bibfield{author}{\bibinfo{person}{Roy Assaf} {and} \bibinfo{person}{Anika
  Schumann}.} \bibinfo{year}{2019}\natexlab{}.
\newblock \showarticletitle{Explainable Deep Neural Networks for Multivariate
  Time Series Predictions.}. In \bibinfo{booktitle}{\emph{IJCAI}}.
  \bibinfo{pages}{6488--6490}.
\newblock


\bibitem[\protect\citeauthoryear{Bach, Binder, Montavon, Klauschen, M{\"u}ller,
  and Samek}{Bach et~al\mbox{.}}{2015}]%
        {bach2015pixel}
\bibfield{author}{\bibinfo{person}{Sebastian Bach}, \bibinfo{person}{Alexander
  Binder}, \bibinfo{person}{Gr{\'e}goire Montavon}, \bibinfo{person}{Frederick
  Klauschen}, \bibinfo{person}{Klaus-Robert M{\"u}ller}, {and}
  \bibinfo{person}{Wojciech Samek}.} \bibinfo{year}{2015}\natexlab{}.
\newblock \showarticletitle{On pixel-wise explanations for non-linear
  classifier decisions by layer-wise relevance propagation}.
\newblock \bibinfo{journal}{\emph{PloS one}} \bibinfo{volume}{10},
  \bibinfo{number}{7} (\bibinfo{year}{2015}), \bibinfo{pages}{e0130140}.
\newblock


\bibitem[\protect\citeauthoryear{Bhatt, Weller, and Moura}{Bhatt
  et~al\mbox{.}}{2020}]%
        {ijcai2020-417}
\bibfield{author}{\bibinfo{person}{Umang Bhatt}, \bibinfo{person}{Adrian
  Weller}, {and} \bibinfo{person}{José M.~F. Moura}.}
  \bibinfo{year}{2020}\natexlab{}.
\newblock \showarticletitle{Evaluating and Aggregating Feature-based Model
  Explanations}. In \bibinfo{booktitle}{\emph{Proceedings of the Twenty-Ninth
  International Joint Conference on Artificial Intelligence, {IJCAI-20}}},
  \bibfield{editor}{\bibinfo{person}{Christian Bessiere}} (Ed.).
  \bibinfo{publisher}{International Joint Conferences on Artificial
  Intelligence Organization}, \bibinfo{pages}{3016--3022}.
\newblock
\urldef\tempurl%
\url{https://doi.org/10.24963/ijcai.2020/417}
\showDOI{\tempurl}
\newblock
\shownote{Main track}.


\bibitem[\protect\citeauthoryear{B{\"u}hlmann}{B{\"u}hlmann}{2002}]%
        {buhlmann2002bootstraps}
\bibfield{author}{\bibinfo{person}{Peter B{\"u}hlmann}.}
  \bibinfo{year}{2002}\natexlab{}.
\newblock \showarticletitle{Bootstraps for time series}.
\newblock \bibinfo{journal}{\emph{Statistical science}} (\bibinfo{year}{2002}),
  \bibinfo{pages}{52--72}.
\newblock


\bibitem[\protect\citeauthoryear{Chen and Guestrin}{Chen and Guestrin}{2016}]%
        {xgboost_paper}
\bibfield{author}{\bibinfo{person}{Tianqi Chen} {and} \bibinfo{person}{Carlos
  Guestrin}.} \bibinfo{year}{2016}\natexlab{}.
\newblock \showarticletitle{{XGBoost: A Scalable Tree Boosting System}}. In
  \bibinfo{booktitle}{\emph{Proceedings of the 22nd ACM SIGKDD International
  Conference on Knowledge Discovery and Data Mining}}.
  \bibinfo{pages}{785–794}.
\newblock
\urldef\tempurl%
\url{https://doi.org/10.1145/2939672.2939785}
\showDOI{\tempurl}


\bibitem[\protect\citeauthoryear{Efron}{Efron}{1992}]%
        {efron1992bootstrap}
\bibfield{author}{\bibinfo{person}{Bradley Efron}.}
  \bibinfo{year}{1992}\natexlab{}.
\newblock \showarticletitle{Bootstrap methods: another look at the jackknife}.
\newblock In \bibinfo{booktitle}{\emph{Breakthroughs in statistics}}.
  \bibinfo{publisher}{Springer}, \bibinfo{pages}{569--593}.
\newblock


\bibitem[\protect\citeauthoryear{Garc{\'\i}a and Aznarte}{Garc{\'\i}a and
  Aznarte}{2020}]%
        {garcia2020shapley}
\bibfield{author}{\bibinfo{person}{Mar{\'\i}a~Vega Garc{\'\i}a} {and}
  \bibinfo{person}{Jos{\'e}~L Aznarte}.} \bibinfo{year}{2020}\natexlab{}.
\newblock \showarticletitle{Shapley additive explanations for NO2 forecasting}.
\newblock \bibinfo{journal}{\emph{Ecological Informatics}}
  \bibinfo{volume}{56} (\bibinfo{year}{2020}), \bibinfo{pages}{101039}.
\newblock


\bibitem[\protect\citeauthoryear{Hyndman and Athanasopoulos}{Hyndman and
  Athanasopoulos}{2018}]%
        {hyndman_book}
\bibfield{author}{\bibinfo{person}{{Robin John} Hyndman} {and}
  \bibinfo{person}{George Athanasopoulos}.} \bibinfo{year}{2018}\natexlab{}.
\newblock \bibinfo{booktitle}{\emph{Forecasting: Principles and Practice}
  (\bibinfo{edition}{2nd} ed.)}.
\newblock \bibinfo{publisher}{OTexts}, \bibinfo{address}{Australia}.
\newblock


\bibitem[\protect\citeauthoryear{Hyndman and Koehler}{Hyndman and
  Koehler}{2006}]%
        {mase_paper}
\bibfield{author}{\bibinfo{person}{Rob~J. Hyndman} {and}
  \bibinfo{person}{Anne~B. Koehler}.} \bibinfo{year}{2006}\natexlab{}.
\newblock \showarticletitle{Another look at measures of forecast accuracy}.
\newblock \bibinfo{journal}{\emph{International Journal of Forecasting}}
  \bibinfo{volume}{22} (\bibinfo{year}{2006}).
\newblock


\bibitem[\protect\citeauthoryear{Ke, Meng, Finley, Wang, Chen, Ma, Ye, and
  Liu}{Ke et~al\mbox{.}}{2017}]%
        {lightgbm_paper}
\bibfield{author}{\bibinfo{person}{Guolin Ke}, \bibinfo{person}{Qi Meng},
  \bibinfo{person}{Thomas Finley}, \bibinfo{person}{Taifeng Wang},
  \bibinfo{person}{Wei Chen}, \bibinfo{person}{Weidong Ma},
  \bibinfo{person}{Qiwei Ye}, {and} \bibinfo{person}{Tie-Yan Liu}.}
  \bibinfo{year}{2017}\natexlab{}.
\newblock \showarticletitle{{LightGBM: A Highly Efficient Gradient Boosting
  Decision Tree}}. In \bibinfo{booktitle}{\emph{Proceedings of the 31st
  International Conference on Neural Information Processing Systems}}.
  \bibinfo{pages}{3149–3157}.
\newblock


\bibitem[\protect\citeauthoryear{Kreiss and Lahiri}{Kreiss and Lahiri}{2012}]%
        {kreiss2012bootstrap}
\bibfield{author}{\bibinfo{person}{Jens-Peter Kreiss} {and}
  \bibinfo{person}{Soumendra~Nath Lahiri}.} \bibinfo{year}{2012}\natexlab{}.
\newblock \showarticletitle{Bootstrap methods for time series}.
\newblock In \bibinfo{booktitle}{\emph{Handbook of statistics}}.
  Vol.~\bibinfo{volume}{30}. \bibinfo{publisher}{Elsevier},
  \bibinfo{pages}{3--26}.
\newblock


\bibitem[\protect\citeauthoryear{Lipovetsky and Conklin}{Lipovetsky and
  Conklin}{2001}]%
        {lipovetsky2001analysis}
\bibfield{author}{\bibinfo{person}{Stan Lipovetsky} {and}
  \bibinfo{person}{Michael Conklin}.} \bibinfo{year}{2001}\natexlab{}.
\newblock \showarticletitle{Analysis of regression in game theory approach}.
\newblock \bibinfo{journal}{\emph{Applied Stochastic Models in Business and
  Industry}} \bibinfo{volume}{17}, \bibinfo{number}{4} (\bibinfo{year}{2001}),
  \bibinfo{pages}{319--330}.
\newblock


\bibitem[\protect\citeauthoryear{Liu, Zeng, Xu, Lai, and Xu}{Liu
  et~al\mbox{.}}{2021}]%
        {liu2021time}
\bibfield{author}{\bibinfo{person}{Minhao Liu}, \bibinfo{person}{Ailing Zeng},
  \bibinfo{person}{Zhijian Xu}, \bibinfo{person}{Qiuxia Lai}, {and}
  \bibinfo{person}{Qiang Xu}.} \bibinfo{year}{2021}\natexlab{}.
\newblock \showarticletitle{Time Series is a Special Sequence: Forecasting with
  Sample Convolution and Interaction}.
\newblock \bibinfo{journal}{\emph{arXiv preprint arXiv:2106.09305}}
  (\bibinfo{year}{2021}).
\newblock


\bibitem[\protect\citeauthoryear{Lundberg, Erion, Chen, DeGrave, Prutkin, Nair,
  Katz, Himmelfarb, Bansal, and Lee}{Lundberg et~al\mbox{.}}{2020}]%
        {treeshap_paper}
\bibfield{author}{\bibinfo{person}{Scott~M. Lundberg}, \bibinfo{person}{Gabriel
  Erion}, \bibinfo{person}{Hugh Chen}, \bibinfo{person}{Alex DeGrave},
  \bibinfo{person}{Jordan~M Prutkin}, \bibinfo{person}{Bala Nair},
  \bibinfo{person}{Ronit Katz}, \bibinfo{person}{Jonathan Himmelfarb},
  \bibinfo{person}{Nisha Bansal}, {and} \bibinfo{person}{Su-In Lee}.}
  \bibinfo{year}{2020}\natexlab{}.
\newblock \showarticletitle{From local explanations to global understanding
  with explainable AI for trees}.
\newblock \bibinfo{journal}{\emph{Nature Machine Intelligence}}
  \bibinfo{volume}{2} (\bibinfo{year}{2020}), \bibinfo{pages}{56--67}.
\newblock
\urldef\tempurl%
\url{https://doi.org/10.1038/s42256-019-0138-9}
\showDOI{\tempurl}


\bibitem[\protect\citeauthoryear{Lundberg and Lee}{Lundberg and Lee}{2017}]%
        {lundberg2017unified}
\bibfield{author}{\bibinfo{person}{Scott~M Lundberg} {and}
  \bibinfo{person}{Su-In Lee}.} \bibinfo{year}{2017}\natexlab{}.
\newblock \showarticletitle{A unified approach to interpreting model
  predictions}. In \bibinfo{booktitle}{\emph{Proceedings of the 31st
  international conference on neural information processing systems}}.
  \bibinfo{pages}{4768--4777}.
\newblock


\bibitem[\protect\citeauthoryear{Madhikermi, Malhi, and
  Fr{\"a}mling}{Madhikermi et~al\mbox{.}}{2019}]%
        {madhikermi2019explainable}
\bibfield{author}{\bibinfo{person}{Manik Madhikermi},
  \bibinfo{person}{Avleen~Kaur Malhi}, {and} \bibinfo{person}{Kary
  Fr{\"a}mling}.} \bibinfo{year}{2019}\natexlab{}.
\newblock \showarticletitle{Explainable artificial intelligence based heat
  recycler fault detection in air handling unit}. In
  \bibinfo{booktitle}{\emph{International Workshop on Explainable, Transparent
  Autonomous Agents and Multi-Agent Systems}}. Springer,
  \bibinfo{pages}{110--125}.
\newblock


\bibitem[\protect\citeauthoryear{Makridakis, Spiliotis, and
  Assimakopoulos}{Makridakis et~al\mbox{.}}{2022}]%
        {makridakis2022m5}
\bibfield{author}{\bibinfo{person}{Spyros Makridakis},
  \bibinfo{person}{Evangelos Spiliotis}, {and} \bibinfo{person}{Vassilios
  Assimakopoulos}.} \bibinfo{year}{2022}\natexlab{}.
\newblock \showarticletitle{M5 accuracy competition: Results, findings, and
  conclusions}.
\newblock \bibinfo{journal}{\emph{International Journal of Forecasting}}
  (\bibinfo{year}{2022}).
\newblock


\bibitem[\protect\citeauthoryear{Miller}{Miller}{2019}]%
        {miller_ai_2019}
\bibfield{author}{\bibinfo{person}{Tim Miller}.}
  \bibinfo{year}{2019}\natexlab{}.
\newblock \showarticletitle{Explanation in artificial intelligence: Insights
  from the social sciences}.
\newblock \bibinfo{journal}{\emph{Artificial Intelligence}}
  \bibinfo{volume}{267} (\bibinfo{year}{2019}), \bibinfo{pages}{1--38}.
\newblock
\urldef\tempurl%
\url{https://doi.org/10.1016/j.artint.2018.07.007}
\showDOI{\tempurl}


\bibitem[\protect\citeauthoryear{Mokhtari, Higdon, and Ba{\c{s}}ar}{Mokhtari
  et~al\mbox{.}}{2019}]%
        {mokhtari2019interpreting}
\bibfield{author}{\bibinfo{person}{Karim~El Mokhtari},
  \bibinfo{person}{Ben~Peachey Higdon}, {and} \bibinfo{person}{Ay{\c{s}}e
  Ba{\c{s}}ar}.} \bibinfo{year}{2019}\natexlab{}.
\newblock \showarticletitle{Interpreting financial time series with SHAP
  values}. In \bibinfo{booktitle}{\emph{Proceedings of the 29th Annual
  International Conference on Computer Science and Software Engineering}}.
  \bibinfo{pages}{166--172}.
\newblock


\bibitem[\protect\citeauthoryear{Molnar}{Molnar}{2019}]%
        {molnar_book_2019}
\bibfield{author}{\bibinfo{person}{Christoph Molnar}.}
  \bibinfo{year}{2019}\natexlab{}.
\newblock \bibinfo{booktitle}{\emph{Interpretable Machine Learning}}.
\newblock
\urldef\tempurl%
\url{https://christophm.github.io/interpretable-ml-book}
\showURL{%
\tempurl}


\bibitem[\protect\citeauthoryear{Oreshkin, Carpov, Chapados, and
  Bengio}{Oreshkin et~al\mbox{.}}{2019}]%
        {oreshkin2019n}
\bibfield{author}{\bibinfo{person}{Boris~N Oreshkin}, \bibinfo{person}{Dmitri
  Carpov}, \bibinfo{person}{Nicolas Chapados}, {and} \bibinfo{person}{Yoshua
  Bengio}.} \bibinfo{year}{2019}\natexlab{}.
\newblock \showarticletitle{N-BEATS: Neural basis expansion analysis for
  interpretable time series forecasting}. In
  \bibinfo{booktitle}{\emph{International Conference on Learning
  Representations}}.
\newblock


\bibitem[\protect\citeauthoryear{Prokhorenkova, Gusev, Vorobev, Dorogush, and
  Gulin}{Prokhorenkova et~al\mbox{.}}{2018}]%
        {catboost_paper}
\bibfield{author}{\bibinfo{person}{Liudmila Prokhorenkova},
  \bibinfo{person}{Gleb Gusev}, \bibinfo{person}{Aleksandr Vorobev},
  \bibinfo{person}{Anna~Veronika Dorogush}, {and} \bibinfo{person}{Andrey
  Gulin}.} \bibinfo{year}{2018}\natexlab{}.
\newblock \showarticletitle{{CatBoost: Unbiased Boosting with Categorical
  Features}}. In \bibinfo{booktitle}{\emph{Proceedings of the 32nd
  International Conference on Neural Information Processing Systems}}.
  \bibinfo{pages}{6639–6649}.
\newblock


\bibitem[\protect\citeauthoryear{Rajapaksha, Bergmeir, and Hyndman}{Rajapaksha
  et~al\mbox{.}}{2021}]%
        {DBLP:journals/corr/abs-2111-07001}
\bibfield{author}{\bibinfo{person}{Dilini Rajapaksha},
  \bibinfo{person}{Christoph Bergmeir}, {and} \bibinfo{person}{Rob~J.
  Hyndman}.} \bibinfo{year}{2021}\natexlab{}.
\newblock \showarticletitle{LoMEF: {A} Framework to Produce Local Explanations
  for Global Model Time Series Forecasts}.
\newblock \bibinfo{journal}{\emph{CoRR}}  \bibinfo{volume}{abs/2111.07001}
  (\bibinfo{year}{2021}).
\newblock
\showeprint[arXiv]{2111.07001}
\urldef\tempurl%
\url{https://arxiv.org/abs/2111.07001}
\showURL{%
\tempurl}


\bibitem[\protect\citeauthoryear{Ribeiro, Singh, and Guestrin}{Ribeiro
  et~al\mbox{.}}{2016}]%
        {ribeiro2016should}
\bibfield{author}{\bibinfo{person}{Marco~Tulio Ribeiro},
  \bibinfo{person}{Sameer Singh}, {and} \bibinfo{person}{Carlos Guestrin}.}
  \bibinfo{year}{2016}\natexlab{}.
\newblock \showarticletitle{"Why should i trust you?" Explaining the
  predictions of any classifier}. In \bibinfo{booktitle}{\emph{Proceedings of
  the 22nd ACM SIGKDD international conference on knowledge discovery and data
  mining}}. \bibinfo{pages}{1135--1144}.
\newblock


\bibitem[\protect\citeauthoryear{Salinas, Flunkert, Gasthaus, and
  Januschowski}{Salinas et~al\mbox{.}}{2020}]%
        {salinas2020deepar}
\bibfield{author}{\bibinfo{person}{David Salinas}, \bibinfo{person}{Valentin
  Flunkert}, \bibinfo{person}{Jan Gasthaus}, {and} \bibinfo{person}{Tim
  Januschowski}.} \bibinfo{year}{2020}\natexlab{}.
\newblock \showarticletitle{DeepAR: Probabilistic forecasting with
  autoregressive recurrent networks}.
\newblock \bibinfo{journal}{\emph{International Journal of Forecasting}}
  \bibinfo{volume}{36}, \bibinfo{number}{3} (\bibinfo{year}{2020}),
  \bibinfo{pages}{1181--1191}.
\newblock


\bibitem[\protect\citeauthoryear{Saluja, Malhi, Knapi{\v{c}}, Fr{\"a}mling, and
  Cavdar}{Saluja et~al\mbox{.}}{2021}]%
        {saluja2021towards}
\bibfield{author}{\bibinfo{person}{Rohit Saluja}, \bibinfo{person}{Avleen
  Malhi}, \bibinfo{person}{Samanta Knapi{\v{c}}}, \bibinfo{person}{Kary
  Fr{\"a}mling}, {and} \bibinfo{person}{Cicek Cavdar}.}
  \bibinfo{year}{2021}\natexlab{}.
\newblock \showarticletitle{Towards a Rigorous Evaluation of Explainability for
  Multivariate Time Series}.
\newblock \bibinfo{journal}{\emph{arXiv preprint arXiv:2104.04075}}
  (\bibinfo{year}{2021}).
\newblock


\bibitem[\protect\citeauthoryear{Schlegel, Arnout, El-Assady, Oelke, and
  Keim}{Schlegel et~al\mbox{.}}{2019}]%
        {schlegel2019towards}
\bibfield{author}{\bibinfo{person}{Udo Schlegel}, \bibinfo{person}{Hiba
  Arnout}, \bibinfo{person}{Mennatallah El-Assady}, \bibinfo{person}{Daniela
  Oelke}, {and} \bibinfo{person}{Daniel~A Keim}.}
  \bibinfo{year}{2019}\natexlab{}.
\newblock \showarticletitle{Towards a rigorous evaluation of xai methods on
  time series}. In \bibinfo{booktitle}{\emph{2019 IEEE/CVF International
  Conference on Computer Vision Workshop (ICCVW)}}. IEEE,
  \bibinfo{pages}{4197--4201}.
\newblock


\bibitem[\protect\citeauthoryear{Shrikumar, Greenside, and Kundaje}{Shrikumar
  et~al\mbox{.}}{2017}]%
        {shrikumar2017learning}
\bibfield{author}{\bibinfo{person}{Avanti Shrikumar}, \bibinfo{person}{Peyton
  Greenside}, {and} \bibinfo{person}{Anshul Kundaje}.}
  \bibinfo{year}{2017}\natexlab{}.
\newblock \showarticletitle{Learning important features through propagating
  activation differences}. In \bibinfo{booktitle}{\emph{International
  Conference on Machine Learning}}. PMLR, \bibinfo{pages}{3145--3153}.
\newblock


\bibitem[\protect\citeauthoryear{Smyl}{Smyl}{2020}]%
        {smyl2020hybrid}
\bibfield{author}{\bibinfo{person}{Slawek Smyl}.}
  \bibinfo{year}{2020}\natexlab{}.
\newblock \showarticletitle{A hybrid method of exponential smoothing and
  recurrent neural networks for time series forecasting}.
\newblock \bibinfo{journal}{\emph{International Journal of Forecasting}}
  \bibinfo{volume}{36}, \bibinfo{number}{1} (\bibinfo{year}{2020}),
  \bibinfo{pages}{75--85}.
\newblock


\bibitem[\protect\citeauthoryear{Taylor and Letham}{Taylor and Letham}{2018}]%
        {taylor2018forecasting}
\bibfield{author}{\bibinfo{person}{Sean~J Taylor} {and}
  \bibinfo{person}{Benjamin Letham}.} \bibinfo{year}{2018}\natexlab{}.
\newblock \showarticletitle{Forecasting at scale}.
\newblock \bibinfo{journal}{\emph{The American Statistician}}
  \bibinfo{volume}{72}, \bibinfo{number}{1} (\bibinfo{year}{2018}),
  \bibinfo{pages}{37--45}.
\newblock


\bibitem[\protect\citeauthoryear{Zhou, Zhang, Peng, Zhang, Li, Xiong, and
  Zhang}{Zhou et~al\mbox{.}}{2021}]%
        {zhou2021informer}
\bibfield{author}{\bibinfo{person}{Haoyi Zhou}, \bibinfo{person}{Shanghang
  Zhang}, \bibinfo{person}{Jieqi Peng}, \bibinfo{person}{Shuai Zhang},
  \bibinfo{person}{Jianxin Li}, \bibinfo{person}{Hui Xiong}, {and}
  \bibinfo{person}{Wancai Zhang}.} \bibinfo{year}{2021}\natexlab{}.
\newblock \showarticletitle{Informer: Beyond efficient transformer for long
  sequence time-series forecasting}. In \bibinfo{booktitle}{\emph{Proceedings
  of AAAI}}.
\newblock


\end{thebibliography}

\newpage
\section{Reproducibility}\label{sec:Reproducibility}

\subsection{Data availability}
\begin{enumerate}
    \item jeans-sales-daily: Proprietary dataset.
    \item jeans-sales-weekly: Proprietary dataset.
    \item us-unemployment:\\
    \url{https://www.bls.gov/data/#unemployment}
    \item peyton-manning:\\ \url{https://en.wikipedia.org/wiki/Peyton_Manning}
    \item bike-sharing: \url{http://archive.ics.uci.edu/ml/datasets/Bike+Sharing+Dataset}
\end{enumerate}

\subsection{Detailed results}
We have provided a summarized observation from our experiments in the main text (\S~\ref{sec:Results}). Here, we demonstrate the detailed results obtained in the experiments.
\subsubsection{Detailed Accuracy of the surrogate model}
Table~\ref{tab:metrics-surrogate} show the errors of the surrogate model obtained through time series cross validation as explained in \S~\ref{sec:Experiments}.
\begin{table}[ht]
\centering
\small
\caption{Backtested error metrics for the surrogate model for different forecasters and datasets.}
\renewcommand{\arraystretch}{0.77}
\begin{tabular}{lrrrr}
\toprule
\textbf{forecaster} & \textbf{MAE} & \textbf{RMSE} & \textbf{MAPE} & \textbf{MASE}  \\
\midrule
\multicolumn{5}{l}{\textbf{Naive}}\\
jeans-sales-weekly & 43.69 & 44.78 & 0.01 & 0.10 \\
jeans-sales-daily & 17.09 & 17.17 & 0.02 & 0.15 \\
us-unemployment & 0.00 & 0.00 & 0.00 & 0.02 \\
peyton-manning & 0.06 & 0.06 & 0.01 & 0.19 \\
bike-sharing & 52.43 & 56.78 & 0.01 & 0.08 \\
\midrule
\multicolumn{5}{l}{\textbf{SeasonalNaive}}\\
jeans-sales-weekly & 75.65 & 116.52 & 0.03 & 0.18 \\
jeans-sales-daily & 4.45 & 14.55 & 0.01 & 0.04 \\
us-unemployment & 0.01 & 0.02 & 0.00 & 0.08 \\
peyton-manning & 0.01 & 0.02 & 0.00 & 0.04 \\
bike-sharing & 262.34 & 363.34 & 0.07 & 0.42 \\
\midrule
\multicolumn{5}{l}{\textbf{MovingAverage(k=6)}}\\
jeans-sales-weekly & 347.26 & 359.37 & 0.10 & 0.78 \\
jeans-sales-daily & 61.74 & 62.42 & 0.12 & 0.54 \\
us-unemployment & 0.11 & 0.11 & 0.01 & 0.66 \\
peyton-manning & 0.05 & 0.05 & 0.01 & 0.16 \\
bike-sharing & 158.67 & 161.28 & 0.03 & 0.24 \\
\midrule
\multicolumn{5}{l}{\textbf{SimpleExponentialSmoothing(0.5)}}\\
jeans-sales-weekly & 239.29 & 261.30 & 0.08 & 0.55 \\
jeans-sales-daily & 101.76 & 104.07 & 0.16 & 0.85 \\
us-unemployment & 0.09 & 0.09 & 0.01 & 0.54 \\
peyton-manning & 0.11 & 0.11 & 0.01 & 0.35 \\
bike-sharing & 256.76 & 259.95 & 0.05 & 0.39 \\
\midrule
\multicolumn{5}{l}{\textbf{Prophet}}\\
jeans-sales-weekly & 463.60 & 586.44 & 0.16 & 1.07 \\
jeans-sales-daily & 100.64 & 119.25 & 0.15 & 0.87 \\
us-unemployment & 0.95 & 1.04 & 0.19 & 5.88 \\
peyton-manning & 0.46 & 0.53 & 0.06 & 1.50 \\
bike-sharing & 650.72 & 739.13 & 0.13 & 1.01 \\
\midrule
\multicolumn{5}{l}{\textbf{XGBoost}}\\
jeans-sales-weekly & 398.31 & 485.52 & 0.15 & 0.91 \\
jeans-sales-daily & 97.39 & 123.92 & 0.21 & 0.86 \\
us-unemployment & 0.96 & 1.06 & 0.15 & 6.04 \\
peyton-manning & 0.96 & 1.16 & 0.12 & 3.11 \\
bike-sharing & 879.15 & 981.73 & 0.18 & 1.36 \\
\bottomrule
\end{tabular}
\label{tab:metrics-surrogate}
\end{table}

\subsubsection{Evaluation of TsSHAP explanations}
Table~\ref{tab:metrics-global}, Table~\ref{tab:metrics-semilocal}, and Table~\ref{tab:metrics-local} provide the detailed results of the evaluation of TsSHAP explanations using faithfulness, sensitivity, and complexity metrics for global, semi-local, and local explanations respectively.

\begin{figure*}
\centering
\begin{subfigure}[b]{\columnwidth}
    \centering
    \includegraphics[width=\columnwidth]{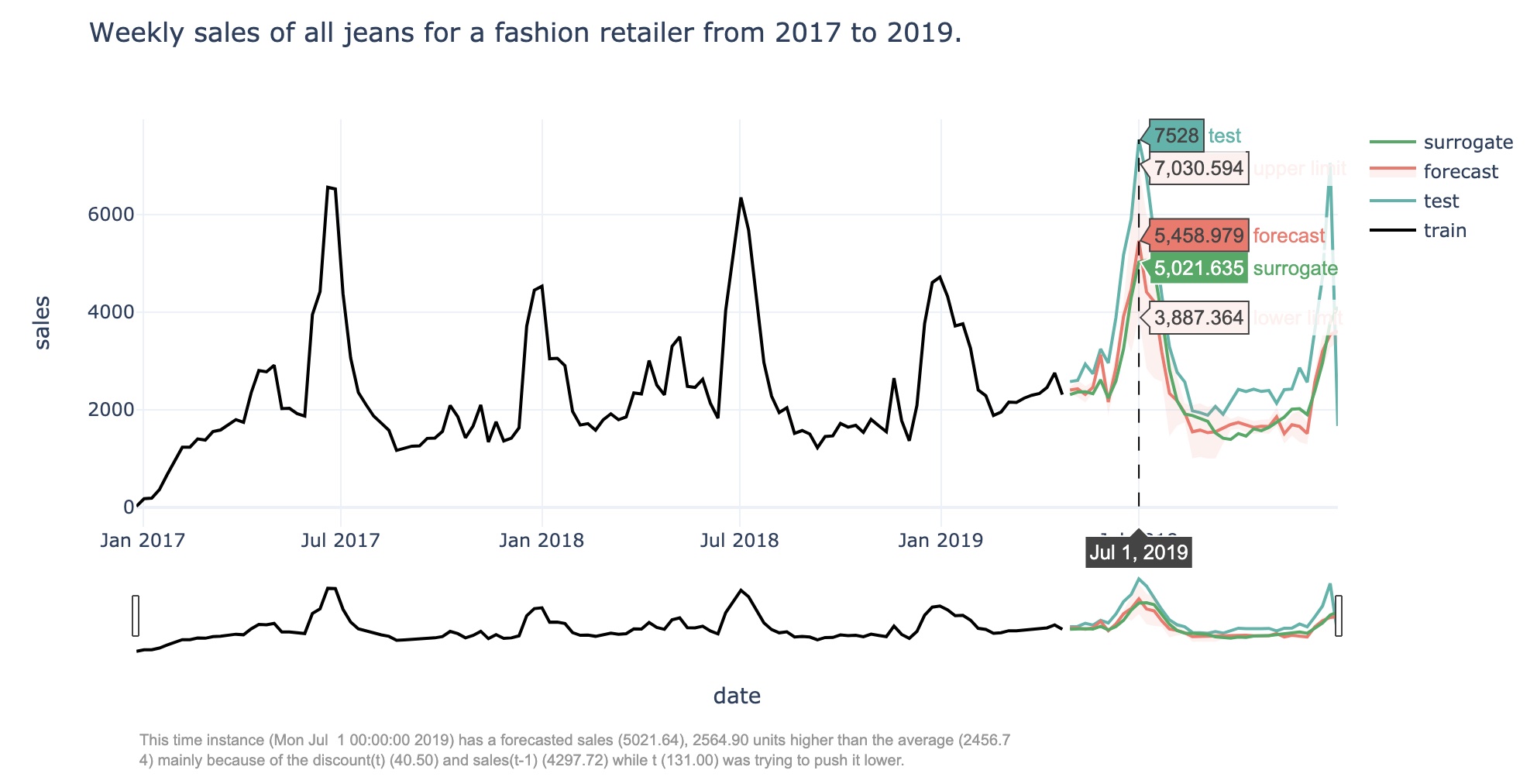}
    \caption{Forecasts from the forecaster and the surrogate model}
\end{subfigure}
\hfill
\begin{subfigure}[b]{\columnwidth}
    \centering
    \includegraphics[width=\columnwidth]{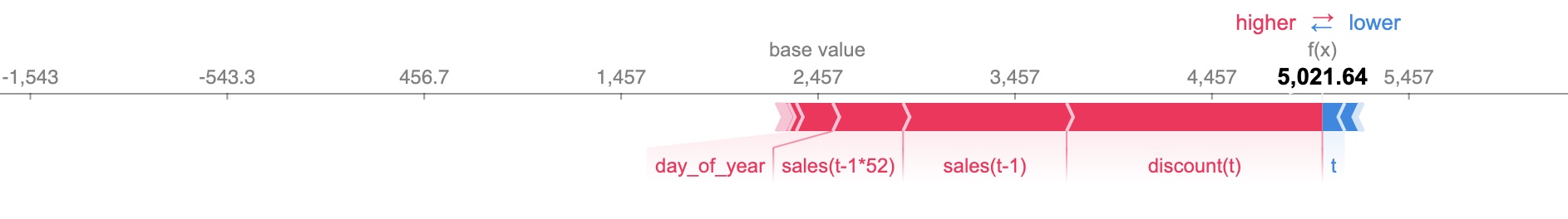}
    \caption{Local explanation}
\end{subfigure}
\hfill
\begin{subfigure}[b]{0.49\columnwidth}
    \centering
    \includegraphics[width=\columnwidth]{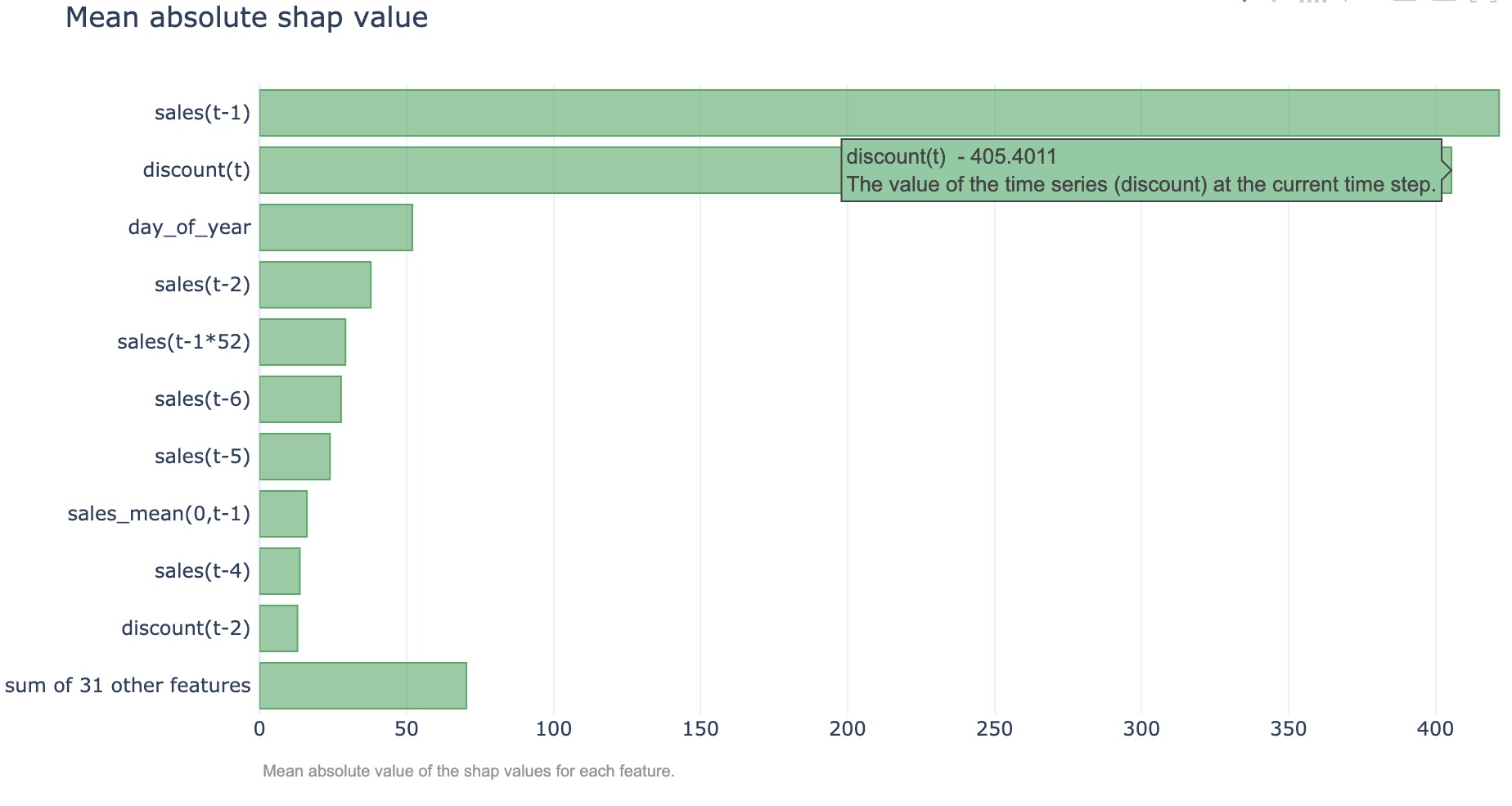}
    \caption{Global explanation}
\end{subfigure}
\hfill
\begin{subfigure}[b]{0.49\columnwidth}
    \centering
    \includegraphics[width=\columnwidth]{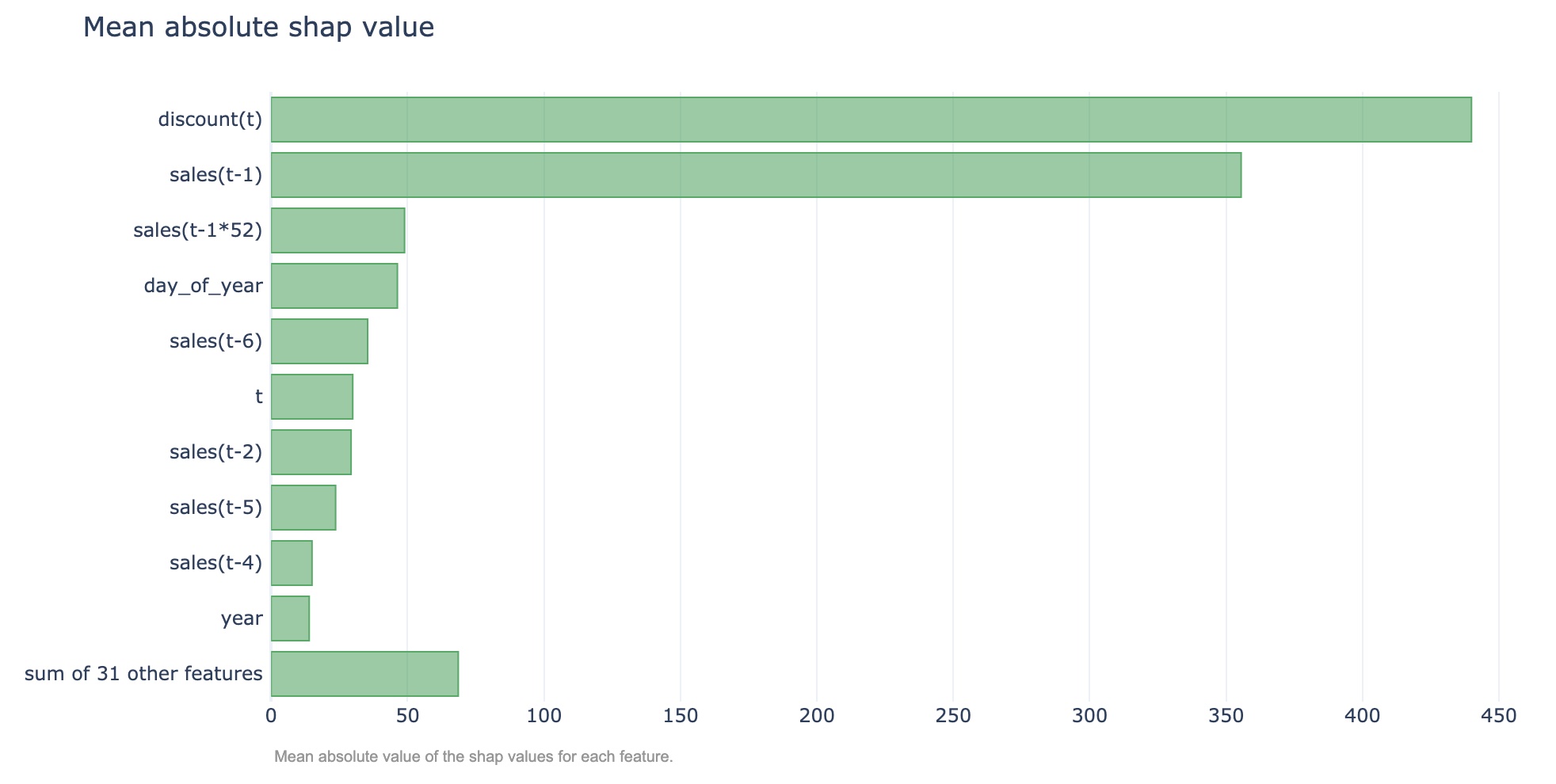}
    \caption{Semi-local explanation}
\end{subfigure} 
\hfill
\begin{subfigure}[b]{0.49\columnwidth}
    \centering
    \includegraphics[width=\columnwidth]{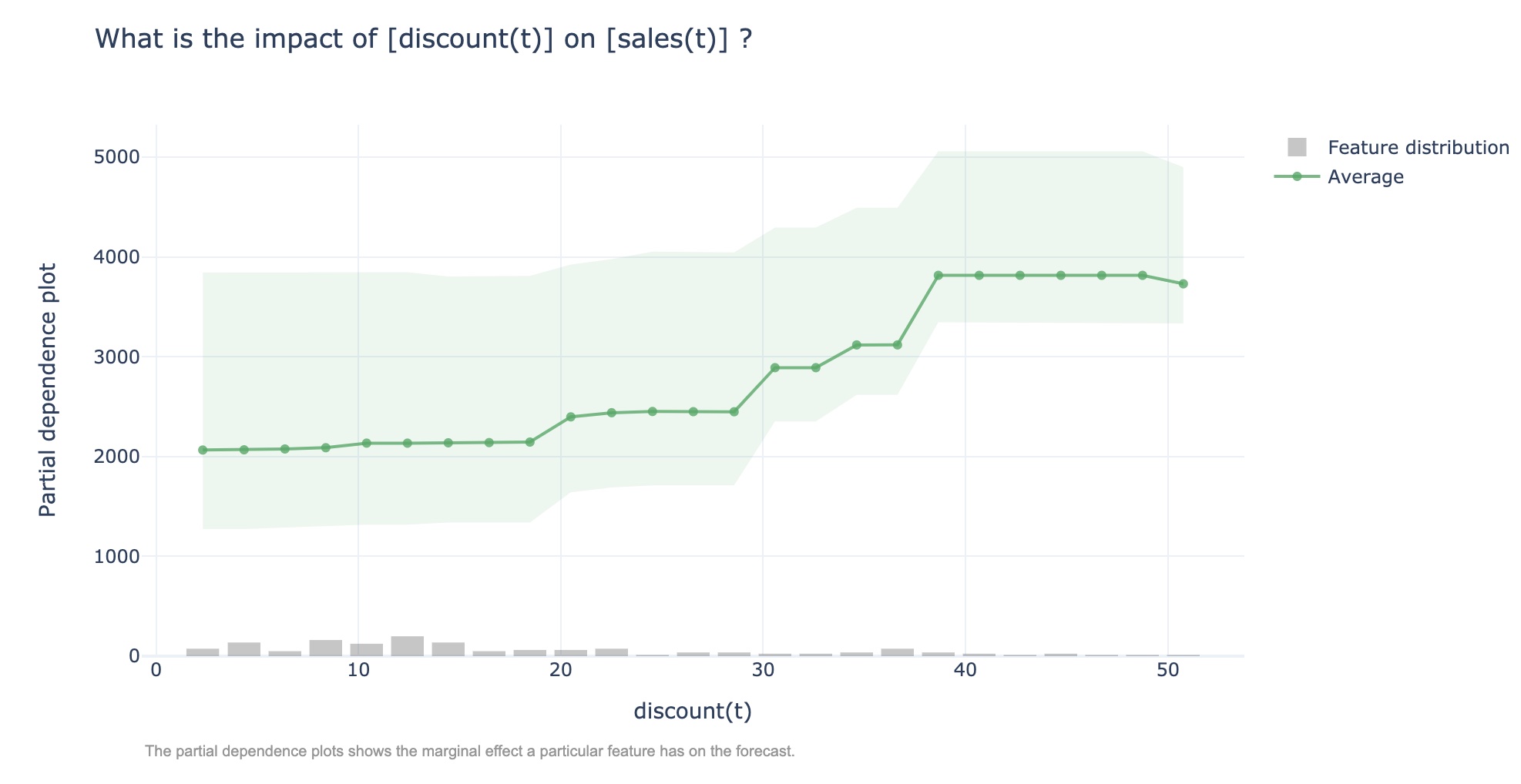}
    \caption{Partial dependence plot}
\end{subfigure}
\hfill
\begin{subfigure}[b]{0.49\columnwidth}
    \centering
    \includegraphics[width=\columnwidth]{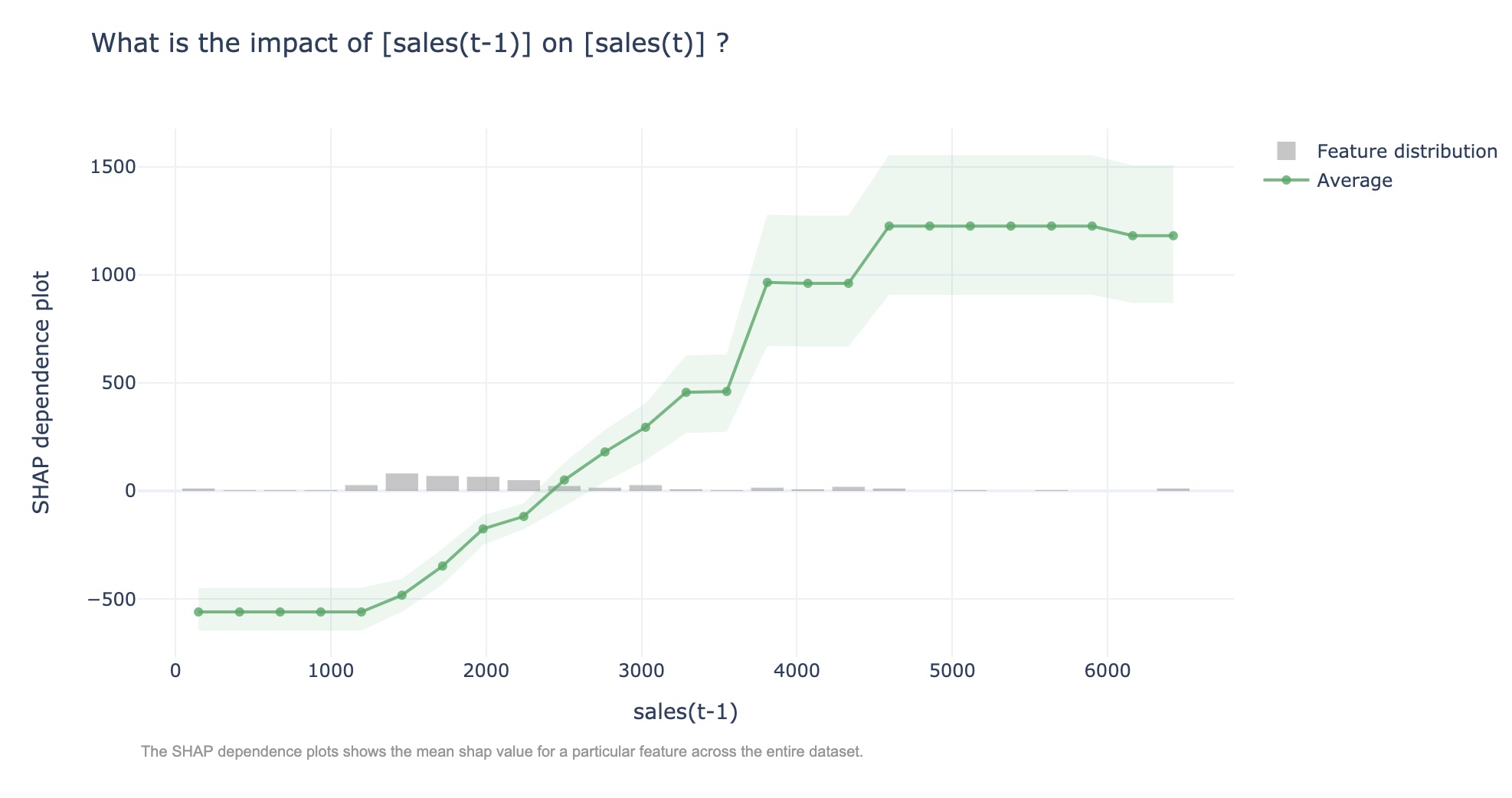}
    \caption{SHAP dependence plot}
 \end{subfigure} 
\caption{Explanations for \textbf{XGBoost} forecaster on the \textbf{jeans-sales-weekly} dataset.} 
\label{fig:XGBoost-jeans-sales-weekly}
\end{figure*}

\subsection{Some more TsSHAP illustrations}
Figure~\ref{fig:XGBoost-jeans-sales-weekly} shows the TsSHAP explanations for XGBoost forecaster on the \texttt{jeans-sales-weekly} data. 
We can see that the surrogate model accurately approximates the forecaster in this case.
Moreover, a closer look into the local explanation says that \texttt{sales(t-1)}, last year's value \texttt{sales(t-52)}, and the external regressor \texttt{discount(t)} are primarily bringing the forecast up from the average sales.
The global explanation shows that \texttt{sales(t-1)} and \texttt{discount(t)} are the main explanatory factors for the forecast made by the forecaster across the entire time series data.
The semi-local explanation depicts similar picture for the entire forecast horizon interval.
The specific PDP shows that after a certain threshold, increasing discount is helpful for increasing sales, but it can also saturate at some point.
The specific SDP shows \texttt{sales(t)} is positively correlated with \texttt{sales(t-1)}.

TsSHAP also provides correct explanations for the Naive forecasters, \ie it picks the last observation as the most important feature in all the scopes.
We skip the illustration for the Naive forecaster due to space limitation.

\begin{table}[H]
\small
\centering
\caption{Metrics for global explanations.}
\renewcommand{\arraystretch}{0.75}
\begin{tabular}{lrrr}
\toprule
& \multicolumn{3}{c}{\small{explanation metrics}}\\
\cmidrule(lr){2-4}
\small{forecaster} & \small{faithfulness} & \small{sensitivity} & \small{complexity}\\
\midrule
\multicolumn{4}{l}{\textbf{Naive}}\\
jeans-sales-weekly & 0.05 & 43.99 & 0.08 \\
jeans-sales-daily & 0.24 & 2.68 & 0.07 \\
us-unemployment & 0.26 & 0.01 & 0.04 \\
peyton-manning & 0.08 & 0.00 & 0.09 \\
bike-sharing & 0.15 & 10.47 & 0.09 \\
\midrule
\multicolumn{4}{l}{\textbf{SeasonalNaive}}\\
jeans-sales-weekly & 0.42 & 65.22 & 0.28 \\
jeans-sales-daily & 0.23 & 51.34 & 0.06 \\
us-unemployment & 0.20 & 0.01 & 0.02 \\
peyton-manning & 0.02 & 0.01 & 0.07 \\
bike-sharing & 0.11 & 448.40 & 0.04 \\
\midrule
\multicolumn{4}{l}{\textbf{MovingAverage(k=6)}}\\
jeans-sales-weekly & 0.45 & 25.58 & 0.23 \\
jeans-sales-daily & 0.20 & 0.80 & 0.08 \\
us-unemployment & 0.35 & 0.02 & 0.24 \\
peyton-manning & 0.06 & 0.00 & 0.13 \\
bike-sharing & 0.16 & 10.68 & 0.13 \\
\midrule
\multicolumn{4}{l}{\textbf{SimpleExponentialSmoothing(0.5)}}\\
jeans-sales-weekly & 0.02 & 62.34 & 1.57 \\
jeans-sales-daily & 0.08 & 25.01 & 1.50 \\
us-unemployment & 0.01 & 0.11 & 1.44 \\
peyton-manning & 0.42 & 0.04 & 1.46 \\
bike-sharing & 0.09 & 79.14 & 1.34 \\
\midrule
\multicolumn{4}{l}{\textbf{Prophet}}\\
jeans-sales-weekly & 0.14 & 130.16 & 2.01 \\
jeans-sales-daily & 0.36 & 38.52 & 2.10 \\
us-unemployment & 0.00 & 0.05 & 1.62 \\
peyton-manning & 0.04 & 0.06 & 2.37 \\
bike-sharing & 0.32 & 154.21 & 2.14 \\
\midrule
\multicolumn{4}{l}{\textbf{XGBoost}}\\
jeans-sales-weekly & 0.00 & 217.22 & 1.75 \\
jeans-sales-daily & 0.36 & 56.39 & 2.06 \\
us-unemployment & 0.22 & 0.07 & 0.85 \\
peyton-manning & 0.06 & 0.20 & 1.95 \\
bike-sharing & 0.001 & 139.71 & 2.67 \\
\bottomrule
\end{tabular}
\label{tab:metrics-global}
\end{table}

\begin{table}[H]
\centering
\small
\caption{Metrics for local explanations.}
\renewcommand{\arraystretch}{0.75}
\begin{tabular}{lrrr}
\toprule
& \multicolumn{3}{c}{\small{explanation metrics}}\\
\cmidrule(lr){2-4}
\small{forecaster} & \small{faithfulness} & \small{sensitivity} & \small{complexity}\\
\midrule
\multicolumn{4}{l}{\textbf{Naive}}\\
jeans-sales-weekly & 0.27 & 174.68 & 0.13 \\
jeans-sales-daily & 0.87 & 13.38 & 0.08 \\
us-unemployment & 0.50 & 0.09 & 0.06 \\
peyton-manning & 0.41 & 0.17 & 0.08 \\
bike-sharing & 0.34 & 435.12 & 0.08 \\
\midrule
\multicolumn{4}{l}{\textbf{SeasonalNaive}}\\
jeans-sales-weekly & 0.18 & 288.36 & 0.16 \\
jeans-sales-daily & 0.20 & 74.42 & 0.12 \\
us-unemployment & 0.17 & 0.11 & 0.08 \\
peyton-manning & 0.18 & 0.25 & 0.23 \\
bike-sharing & 0.19 & 549.87 & 0.29 \\
\midrule
\multicolumn{4}{l}{\textbf{MovingAverage(k=6)}}\\
jeans-sales-weekly & 0.40 & 68.44 & 0.23 \\
jeans-sales-daily & 0.89 & 12.64 & 0.08 \\
us-unemployment & 0.95 & 0.11 & 0.21 \\
peyton-manning & 0.54 & 0.05 & 0.10 \\
bike-sharing & 0.49 & 191.81 & 0.13 \\
\midrule
\multicolumn{4}{l}{\textbf{SimpleExponentialSmoothing(0.5)}}\\
jeans-sales-weekly & 0.20 & 140.37 & 1.58 \\
jeans-sales-daily & 0.66 & 26.50 & 1.49 \\
us-unemployment & 0.95 & 0.22 & 1.43 \\
peyton-manning & 0.84 & 0.11 & 1.37 \\
bike-sharing & 0.27 & 195.08 & 1.28 \\
\midrule
\multicolumn{4}{l}{\textbf{Prophet}}\\
jeans-sales-weekly & 0.25 & 323.18 & 1.82 \\
jeans-sales-daily & 0.36 & 90.15 & 2.03 \\
us-unemployment & 0.62 & 0.18 & 1.41 \\
peyton-manning & 0.59 & 0.37 & 2.22 \\
bike-sharing & 0.48 & 508.43 & 1.86 \\
\midrule
\multicolumn{4}{l}{\textbf{XGBoost}}\\
jeans-sales-weekly & 0.44 & 364.76 & 1.59 \\
jeans-sales-daily & 0.62 & 96.66 & 1.85 \\
us-unemployment & 0.32 & 0.36 & 0.77 \\
peyton-manning & 0.68 & 0.59 & 1.89 \\
bike-sharing & 0.60 & 921.83 & 2.44 \\
\bottomrule
\end{tabular}
\label{tab:metrics-local}
\end{table}

\begin{table}[t]
\centering
\small
\caption{Metrics for semi-local explanations.}
\renewcommand{\arraystretch}{0.75}
\begin{tabular}{lrrr}
\toprule
& \multicolumn{3}{c}{\small{explanation metrics}}\\
\cmidrule(lr){2-4}
\small{forecaster} & \small{faithfulness} & \small{sensitivity} & \small{complexity}\\
\midrule
\multicolumn{4}{l}{\textbf{Naive}}\\
jeans-sales-weekly & 0.27 & 174.43 & 0.14 \\
jeans-sales-daily & 0.18 & 13.30 & 0.08 \\
us-unemployment & 0.51 & 0.08 & 0.06 \\
peyton-manning & 0.42 & 0.17 & 0.08 \\
bike-sharing & 0.34 & 434.51 & 0.08 \\
\midrule
\multicolumn{4}{l}{\textbf{SeasonalNaive}}\\
jeans-sales-weekly & 0.18 & 71.85 & 0.13 \\
jeans-sales-daily & 0.75 & 7.70 & 0.08 \\
us-unemployment & 0.18 & 0.04 & 0.09 \\
peyton-manning & 0.56 & 0.01 & 0.10 \\
bike-sharing & 0.51 & 55.32 & 0.15 \\
\midrule
\multicolumn{4}{l}{\textbf{MovingAverage(k=6)}}\\
jeans-sales-weekly & 0.42 & 61.11 & 0.23 \\
jeans-sales-daily & 0.89 & 12.32 & 0.08 \\
us-unemployment & 0.93 & 0.10 & 0.21 \\
peyton-manning & 0.53 & 0.05 & 0.10 \\
bike-sharing & 0.52 & 186.45 & 0.14 \\
\midrule
\multicolumn{4}{l}{\textbf{SimpleExponentialSmoothing(0.5)}}\\
jeans-sales-weekly & 0.10 & 88.58 & 1.58 \\
jeans-sales-daily & 0.32 & 25.89 & 1.50 \\
us-unemployment & 0.99 & 0.22 & 1.44 \\
peyton-manning & 0.84 & 0.11 & 1.37 \\
bike-sharing & 0.25 & 187.61 & 1.28 \\
\midrule
\multicolumn{4}{l}{\textbf{Prophet}}\\
jeans-sales-weekly & 0.26 & 183.38 & 1.88 \\
jeans-sales-daily & 0.60 & 36.88 & 2.12 \\
us-unemployment & 0.08 & 0.13 & 1.41 \\
peyton-manning & 0.60 & 0.09 & 2.33 \\
bike-sharing & 0.19 & 282.65 & 1.92 \\
\midrule
\multicolumn{4}{l}{\textbf{XGBoost}}\\
jeans-sales-weekly & 0.35 & 167.43 & 1.67 \\
jeans-sales-daily & 0.33 & 54.68 & 1.96 \\
us-unemployment & 0.35 & 0.30 & 0.79 \\
peyton-manning & 0.90 & 0.20 & 1.95 \\
bike-sharing & 0.89 & 249.49 & 2.59 \\
\bottomrule
\end{tabular}
\label{tab:metrics-semilocal}
\end{table}

\end{document}